\documentclass[10pt,twocolumn,letterpaper]{article}

\usepackage{cvpr}              

\usepackage[dvipsnames]{xcolor}

\definecolor{cvprblue}{rgb}{0.21,0.49,0.74}
\usepackage[pagebackref,breaklinks,colorlinks,citecolor=cvprblue]{hyperref}
\usepackage{graphicx}
\usepackage{amsmath}
\usepackage{amssymb}
\usepackage{booktabs}
\usepackage{makecell}
\usepackage{float}
\usepackage{multirow}
\usepackage{url}
\usepackage{amsthm}
\newtheorem{theorem}{Theorem}[section]
\usepackage{subcaption}
\usepackage[linesnumbered,lined,ruled]{algorithm2e}

\usepackage[accsupp]{axessibility}

\usepackage[capitalize]{cleveref}
\crefname{section}{Sec.}{Secs.}
\Crefname{section}{Section}{Sections}
\Crefname{table}{Table}{Tables}
\crefname{table}{Tab.}{Tabs.}

\def\paperID{13797} 
\def\confName{CVPR}
\def\confYear{2024}

\title{Towards Efficient Replay in Federated Incremental Learning}

\author{
Yichen Li\\
Huazhong University of Science\\ and Technology, China\\
{\tt\small ycli0204@hust.edu.cn}
\and
Qunwei Li\\
Ant Group, China\\
{\tt\small qunwei.qw@antgroup.com}
\and
Haozhao Wang\\
Huazhong University of Science\\ and Technology, China\\
{\tt\small hz\_wang@hust.edu.cn}
\and
Ruixuan Li\thanks{Ruixuan Li is the corresponding author.}
\\
Huazhong University of Science\\ and Technology, China\\
{\tt\small rxli@hust.edu.cn}
\and
Wenliang Zhong\\
Ant Group, China\\
{\tt\small yice.zwl@antgroup.com}
\and
Guannan Zhang\\
Ant Group, China\\
{\tt\small zgn138592@antgroup.com}
}
\begin{document}
\maketitle
\begin{abstract}
In Federated Learning (FL), the data in each client is typically assumed fixed or static. However, data often comes in an incremental manner in real-world applications, where the data domain may increase dynamically. 
In this work, we study catastrophic forgetting with data heterogeneity in Federated Incremental Learning (FIL) scenarios where edge clients may lack enough storage space to retain full data. 
We propose to employ a simple, generic framework for FIL named Re-Fed, which can coordinate each client to cache important samples for replay. More specifically, when a new task arrives, each client first caches selected previous samples based on their global and local importance. Then, the client trains the local model with both the cached samples and the samples from the new task.
Theoretically, we analyze the ability of Re-Fed to discover important samples for replay thus alleviating the catastrophic forgetting problem. Moreover, we empirically show that Re-Fed achieves competitive performance compared to state-of-the-art methods.
\end{abstract}    
\section{Introduction}
\label{sec:intro}
Federated learning (FL) is a distributed framework that allows multiple edge clients to learn a unified deep learning
model cooperatively while preserving the data privacy of the local clients \cite{mcmahan2017communication,DBLP:conf/cvpr/WangCW022,DBLP:conf/cvpr/LiXSLLSZ22}. Recently, FL has attracted growing attention and been applied to various fields such as recommendation systems \cite{li2021privacy,du2020federated} and smart healthcare \cite{xu2021federated,nguyen2022federated}.

Typically, FL has been actively studied in a static setting, where the number of training samples does not change over time. However, in a realistic FL application, each client may continue collecting new data. It is difficult to learn new data while retaining previous information in machine learning due to the notorious phenomenon known as catastrophic forgetting \cite{ganin2016domain}, leading to performance degradation on previous tasks. This challenge is further compounded in FL settings, where the data in a client remains inaccessible to other ones and clients lack enough storage to retain full previous samples. 

To address this issue, researchers have studied federated incremental learning (FIL), which enables each client to continuously learn from a local private and incremental task stream \cite{criado2022non,lei2022federated}. The authors in \cite{yoon2021federated} aim to personalize the models for each client by decomposing the model parameters into shared parameters and adaptive parameters, facilitating the transfer of common knowledge across similar tasks among clients. FCIL is proposed in \cite{dong2022federated} which specifically focuses on the federated class-incremental learning scenario and a global model is developed by incorporating additional class-imbalance losses. It is studied in \cite{ijcai2022p0303} to utilize extra distilled data at both the server and client sides, where knowledge distillation is employed to mitigate catastrophic forgetting. FedCIL is proposed in \cite{qi2023better} to learn a generative network and reconstruct past samples for replay, improving the retention of previous information.

\begin{figure*}[t]
\centering
  \includegraphics[width=0.95\linewidth]{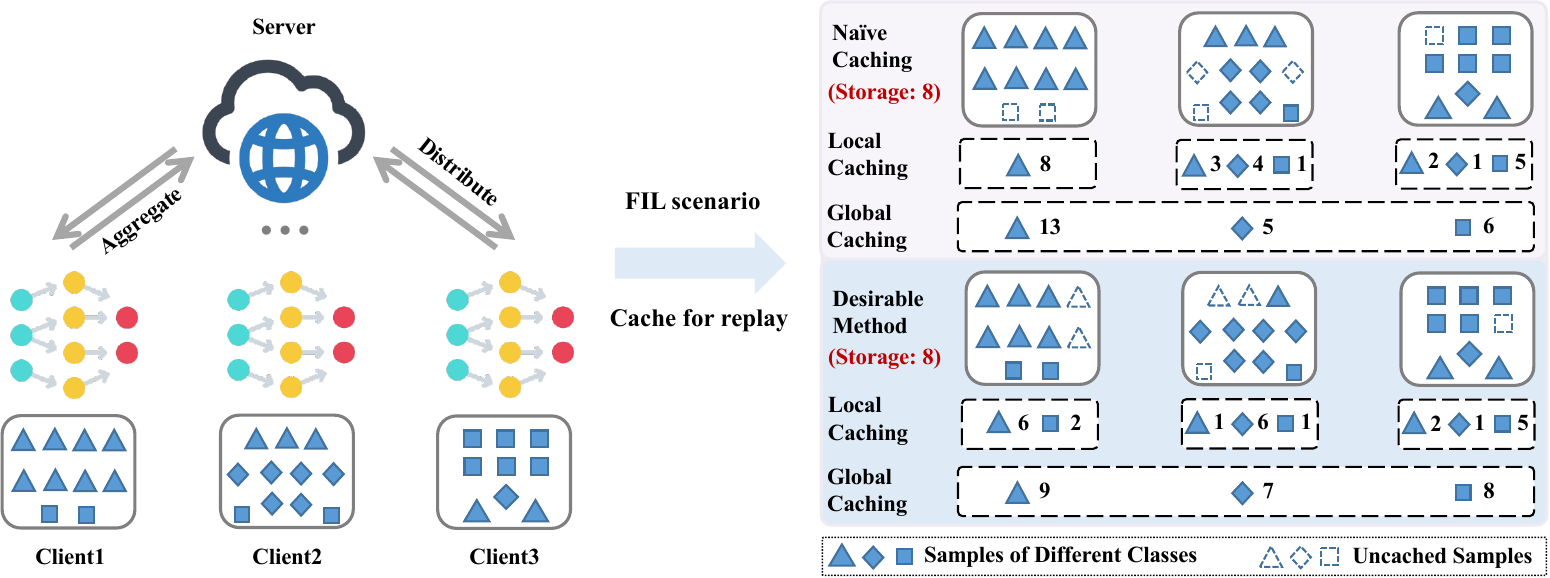}
  \caption{The motivation for our method: an example of 3-client in FIL scenario. When a new task arrives, each client needs to cache previous samples with limited storage for replay, alleviating catastrophic forgetting. Global caching represents all the samples cached by all clients collectively. With a naive caching method, the client may ignore the sample's correlation across clients which increases the statistical data heterogeneity in global caching. With a desirable method, the client tends to cache samples which both considers the distribution of local samples and reduces the statistical data heterogeneity in the global caching.}
  \label{motivation}
  \vspace{-0.2cm}
\end{figure*}
While these approaches may be effective at learning new tasks, it is essential to also consider the constraints of privacy concerns and data heterogeneity. For example, data-reconstruction techniques with gradient or adversarial training are employed in FCIL and FedCIL to alleviate catastrophic forgetting, but it may cause privacy leakage of local data. Moreover, existing works simply assume that each client collects incremental data of different tasks in an independently and identically distributed (IID) manner, ignoring the issue of data heterogeneity in real-world scenarios.

In this paper, we investigate a simple and {efficient} method for catastrophic forgetting in FIL. We consider classification task in FIL and first identify two types of task for newly collected data: (1) Class-Incremental Task: the newly arrived data has different labels from previous data, and thus the label space of data is growing. (2) Domain-Incremental Task: the new data has a domain shift from previous data, and it does not change the label space of data. Then, we assume data heterogeneity as the data of each task in clients is Non-IID.

To tackle the catastrophic forgetting problem in the non-federated environment, data replay \cite{rebuffi2017icarl,mensink2013distance} based methods have demonstrated great effectiveness by caching the important samples from previous tasks and replaying them when learning the new task. 
However, existing replay-based methods fail to consider the correlation between clients in FL, and the cached samples may not be globally optimal for tasks. Essentially, the cached samples should be strongly correlated with the statistical heterogeneity of all data across clients. The intuitive explanation can also be found in a simple 3-client example illustrated in Figure~\ref{motivation}.

To explore this idea, we propose an {efficient} FIL framework named Re-Fed that can alleviate catastrophic forgetting by allowing all clients to synergistically cache data samples.
More specifically, in Re-Fed, each client {caches samples} based not only on their importance in the local dataset but also on their correlation to the global dataset. Here we fist employ an additional personalized informative model (PIM) for each client which can incorporate knowledge of data from global point of view in local caching such that the cached samples contribute to both local and global understanding of the data. Then, we quantify the sample importance by calculating the gradient norm of previous local samples during the update of the PIM. Finally, the client caches the samples with higher importance scores, and trains the local model with both the cached samples from previous tasks and the samples from the new task.

Through extensive experiments on various datasets and two types of newly collected data (Class-Incremental Task and Domain-Incremental Task), we show that Re-Fed significantly improves the model accuracy compared to state-of-the-art approaches. 
The major contributions of this paper are summarized as follows:
\begin{itemize}
    \item We are the first to study the problem of catastrophic forgetting with data heterogeneity in FIL. To address this problem, we propose a novel framework named Re-Fed which can be seen as an off-the-shelf personalization add-on for standard FIL and it inherits privacy protection and efficiency properties as traditional FL applications in FIL scenarios.
    
    \item {Next, we theoretically show that Re-Fed can efficiently discover the important samples for data replay, with guaranteed convergence.}

    \item Finally, we carry out extensive experiments on various datasets and different FIL task scenarios. Experimental results illustrate that our proposed model outperforms the state-of-the-art methods by up to 19.73\% in terms of final accuracy on different tasks.  
\end{itemize}

\section{Related Work}
\label{sec:formatting}
\textbf{Federated Learning}
FL is a technique to train a shared global model by aggregating models from multiple clients that are trained on their own local private datasets \cite{mcmahan2017communication,DBLP:conf/cvpr/WangCW022,DBLP:conf/cvpr/LiXSLLSZ22}. One effective architecture for FL is FedAvg \cite{mcmahan2017communication}, which optimizes the global model by aggregating the parameters of local models trained on private local data. However, traditional FL algorithms like FedAvg face challenges due to data heterogeneity, where the datasets in clients are {Non-IID}, resulting in degradation in model performance \cite{Jeong2018CommunicationEfficientOM,Liu2019edgeAssistedHF}. To tackle the Non-IID issue in FL, {a proximal term is introduced in optimization in \cite{li2020federated} to mitigate the effects of heterogeneous and Non-IID data distribution across participating devices.} Another approach of federated distillation \cite{DBLP:journals/corr/abs-1811-11479}, aims to distill the knowledge from multiple local models into the global model by aggregating only the soft predictions generated by each model. The authors in \cite{lin2020ensemble} proposed a knowledge distillation method that utilizes unlabeled training samples as a proxy dataset. Recently, there has been growing interest in data-free knowledge distillation methods that leverage adversarial approaches in generating data \cite{zhu2021data,DBLP:conf/cvpr/ZhangS0TD22,wang2023dafkd}. However, the aforementioned methods follow a framework designed to handle Non-IID static data with spatial heterogeneity, overlooking the potential challenges posed by incremental tasks with temporal heterogeneity in FL scenarios.

\textbf{Incremental Learning}
Incremental Learning (IL) is a machine learning technique that allows a model to learn continuously from a incremental sequence of tasks while retaining knowledge gained from previous tasks \cite{Hsu2018ReevaluatingCL,vandeVen2019ThreeSF}, including task-incremental learning \cite{Dantam2016IncrementalTA,Maltoni2018ContinuousLI}, class-incremental learning \cite{rebuffi2017icarl,Yu2020SemanticDC}, and domain-incremental learning \cite{mirza2022efficient,Churamani2021DomainIncrementalCL}. Existing approaches in IL can be classified into three main categories: replay-based methods \cite{rebuffi2017icarl,liu2020generative}, regularization-based methods \cite{Jung2020ContinualLW,Yin2020SOLACL}, and parameter isolation methods \cite{Long2015LearningTF,Fernando2017PathNetEC}.
Replay-based methods select representative old samples to retain previously learned knowledge when training on a new task. Regularization-based methods protect existing knowledge from being overwritten with new knowledge by imposing constraints on the loss function of new tasks. Parameter isolation methods typically introduce additional parameters and computations {to learn new tasks}. Here we focus on the federated incremental learning scenario, which can be viewed as a combination of federal learning and incremental learning.


\section{Methodology}
We first formulate two FIL scenarios and propose a simple and scalable framework Re-Fed. Then, we present a scalable algorithm and provide rigorous analytical results to show the efficiency of the proposed method.

\subsection{Problem Formulation}
In the standard IL (non-federated environment), a model learns from a sequence of streaming tasks \{$\mathcal{T}^{1},\mathcal{T}^{2},\cdots, \mathcal{T}^{n}$\} where $\mathcal{T}^{t}$ denotes the $t$-th task of the dataset. Here $\mathcal{T}^{t} = \sum_{i=1}^{N^t}(x^{(i)}_t,y^{(i)}_t)$, which has $N^t$ pairs of sample data $x^{(i)}_t \in \mathcal{X}^{t}$ and corresponding label $y^{(i)}_t \in \mathcal{Y}^{t}$. We use $\mathcal{X}^{t}$ and $\mathcal{Y}^{t}$ to represent the domain space and label space for the $t$-th task, which has $|\mathcal{Y}^{t}|$ classes and $\mathcal{Y}$ = $\bigcup_{t=1}^{n}\mathcal{Y}^{t}$ where $\mathcal{Y}$ denotes the total classes of all time. Similarly, we use $\mathcal{X}$ = $\bigcup_{t=1}^{n}\mathcal{X}^{t}$ to denote the total domain space for tasks of all time. 
In this paper, we focus on two types of IL scenarios: (1) Class-Incremental Task: all tasks share the same domain space, i.e., $\mathcal{X}^{1} = \mathcal{X}^{t}, \forall t \in [n]$. As the sequence of learning tasks arrives, the number of the classes may change, i.e, $\mathcal{Y}^{1} \neq \mathcal{Y}^{t}, \forall t \in [n]$. (2) Domain-Incremental Task: all tasks share the same number of classes i.e., $ \mathcal{Y}^{1} = \mathcal{Y}^{t}, \forall t \in [n]$. As the sequence of tasks arrives, the client needs to learn the new task while their domain and data distribution changes, i.e., $\mathcal{X}^{1} \neq \mathcal{X}^{t}, \forall t \in [n]$.

We further consider IL in a federated setting. We aim to train a global model for $K$ total clients and assume that client $k$ can only access the local private streaming tasks \{$\mathcal{T}_k^{1},\mathcal{T}_k^{2},\cdots, \mathcal{T}_k^{n}$\}. When the $t$-th task comes, while clients can cache all samples from previous tasks without forgetting, the goal is to train a global model $w^t$ over all $t$ tasks $\mathcal{T}^t = \{\sum_{n=1}^t\sum_{k=1}^K \mathcal{T}_k^{n}\}$, which can be formulated as :
\begin{align}\label{parameterAvg}
   w^t= \arg \min_w \sum_{n=1}^t\sum_{k=1}^K\sum_{i=1}^{N^n_k}\frac{1}{|\mathcal{T}^t|} l\left(f_{w_k}(x_{k,n}^{(i)}),y_{k,n}^{(i)}\right). 
\end{align}
where $f_{w_k}(\cdot)$ is the output of the model $w_k$ in client $k$ and $l(\cdot)$ is the cross-entropy loss. Then, due to poor storage in common edge devices, each client caches partial samples for replay. Here we assume each client can only store total $M$ samples and has to cache $M-N^t_k$ samples from $(t-1)$ previous tasks, which is denoted as $\mathcal{T}^{t-1}_{k,cached} = \sum_{i=1}^{M-N^t_k}(\bar{x}_{k,t-1}^{(i)},\bar{y}_{k,t-1}^{(i)})$. The goal is to train a global model $w^t$ over both cached samples and the $t$-th new task, which can be formulated as:
\begin{align}
\label{parameterAvg2}
\begin{split}
   w^t= 
   %
   %
   \arg \min_w \sum_{k=1}^K\sum_{i=1}^{M}\frac{1}{|\mathcal{T}^{t}_{k,local}|} l\left(f_{w_k}(\Tilde{x}_{k,t}^{(i)}),\Tilde{y}_{k,t}^{(i)}\right).
\end{split}
\end{align}
where $\mathcal{T}^{t}_{k,local} = \mathcal{T}^{t-1}_{k,cached}+\mathcal{T}^{t}_{k} = \sum_{i=1}^{M}(\Tilde{x}^{(i)}_{k,t},\Tilde{y}^{(i)}_{k,t})$.

\vspace{5pt}
\subsection{Re-Fed: Framework for FIL}
\begin{figure}[t]
\centering
  \includegraphics[width=\linewidth]{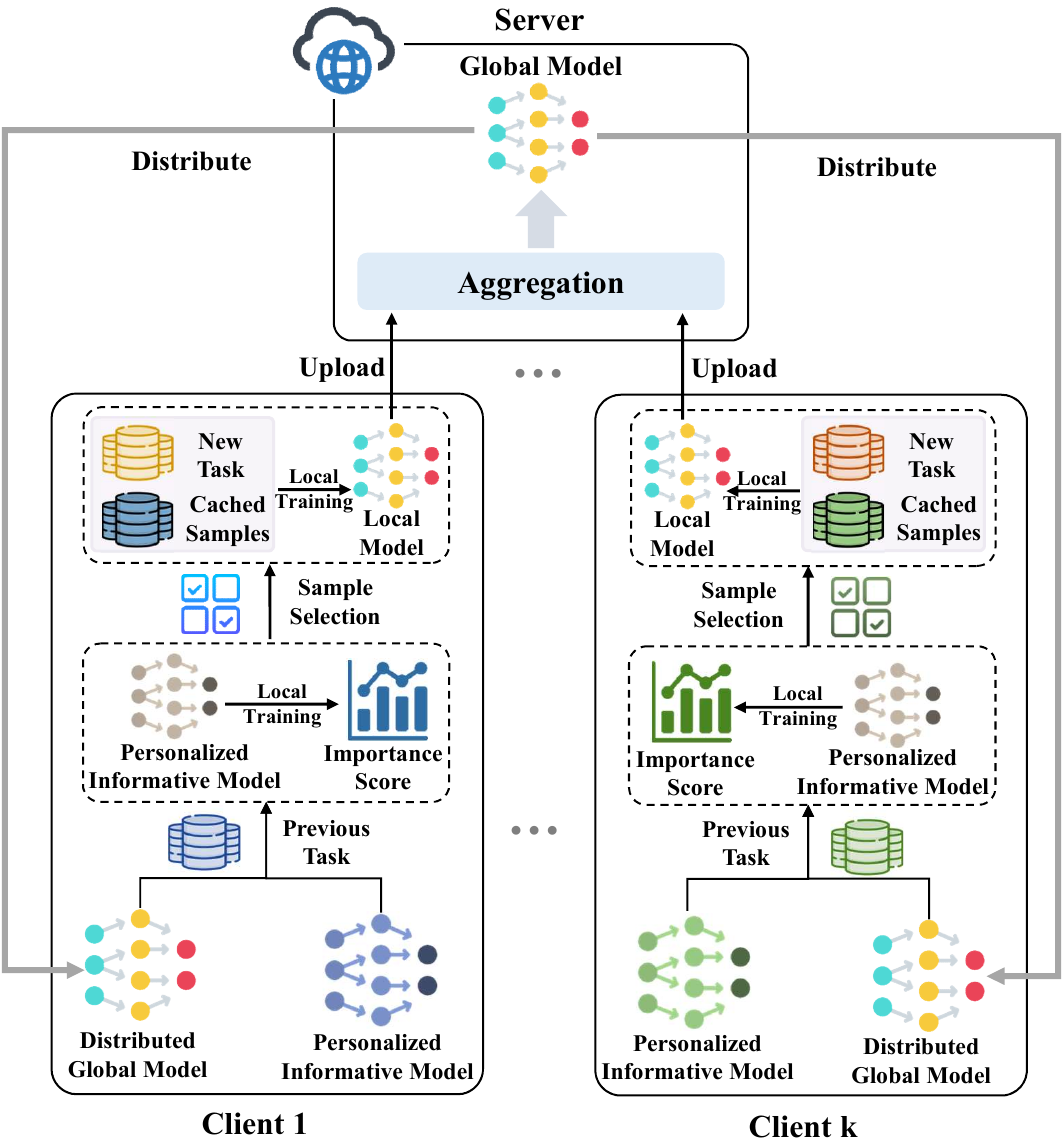}
  \caption{Illustration of the Re-Fed framework. When a new task arrives, each client first updates the personalized informative model on previous local samples with the distributed global model. Then, samples are selected to be cached by the sample importance scores that are calculated with the updated personalized informative model. Finally, each client trains the local model with both the new task and cached previous samples.}
  \label{framework}
\vspace{-0.4cm}
\end{figure}
The key idea of Re-Fed is to identify the sample importance and coordinate clients to cache important previous samples with limited local storage when the new task arrives. Specifically, in each communication round, the clients train the local models with the private sequence of tasks and the server aggregates local models from all participated clients. Then, when new tasks come, each client first trains an extra personalized informative model on previous local samples with the regulation of both the global model and local model. During the update of such a model, gradient norms of individual samples are recorded to calculate sample importance scores. Finally, each client caches samples with higher importance scores based on local storage and continues training the local model on the cached samples and the new task. The workflow of the proposed framework is shown in Algorithm \ref{simplified} and Figure \ref{framework} illustrates the Re-Fed framework. 
\begin{algorithm}[t]
   \caption{Re-Fed}
   \label{simplified}
   \SetKwInOut{Input}{Input}
    \Input{$T$: communication round;\ \ $K$: client number;\ \ $\eta$: learning rate;\ \ $\{\mathcal{T}^{t}\}_{t=1}^n$: distributed dataset with $n$ tasks;\ \ $w$: parameter of the model;\ \ $v_k$: personalized informative model in client $k$.} 
Initialize the parameter $w$; 

    \For(\tcp*[f]{the $t$-th new task}){$c=1$ {\bfseries to} $T$}{
         \ \ Server randomly selects a subset of devices $S_t$ and send $w^{t-1}$ \\      
        \For{each selected client $k \in S_t$ {\rm\bfseries in parallel}}{        
            $\text{Update}\ v_k^{t-1} \text{in $s$ local iterations}$ with (\ref{personalizedmodel}).\\
            
           \For{During the update of $v_k^{t-1}$}{
                Calculate the importance score after total $s$ iterations for the sample $(\Tilde{x}^{(i)}_{k,t-1},\Tilde{y}^{(i)}_{k,t-1})$ with (\ref{importanceScore2}).\\
           }
        Cache previous samples with higher importance scores;\\
        Training the local model with cached samples and the new task with (\ref{LocalTraining});\\
Send the model $w_k^t$ back to the server.
        }
        $w^t \leftarrow \text{ServerAggregation}(\{w_k^t\}_{k \in S_t})$ 
    }
\end{algorithm}

\begin{table*}[t]
    \renewcommand\arraystretch{1.25}
    \centering
    \caption{Performance comparison of various methods in two incremental learning scenarios.}
      \resizebox{\linewidth}{!}{
    \begin{tabular}{c| c| c c |c c c |c c |c}
     \toprule[1pt]
     Scenario & Dataset & FedAvg & FedProx & Fixed & DANN+FL & Shared & FCIL & FedCIL & Re-Fed\\ \hline
    \multirow{3}{*}{Class-Incremental}& CIFAR10 ($\alpha = 1.0$) & 26.73\scriptsize{±1.12} & 25.87\scriptsize{±0.68} & 19.21\scriptsize{±0.06} & 24.86\scriptsize{±2.31} & 23.91\scriptsize{±1.70} & 25.04\scriptsize{±0.11} & 27.35\scriptsize{±1.24} &\textbf{29.22\scriptsize{±0.49}}\\
    & CIFAR100 ($\alpha = 5.0$) & 17.21\scriptsize{±1.35} & 18.03\scriptsize{±0.91} & 9.27\scriptsize{±0.22} & 19.73\scriptsize{±2.17} &18.30\scriptsize{±1.53} & 23.02\scriptsize{±0.66} & 17.98\scriptsize{±1.46} &\textbf{25.61\scriptsize{±0.88}}\\
     & Tiny-ImageNet ($\alpha = 10$) & 27.58\scriptsize{±0.74} & 21.82\scriptsize{±0.90} & 12.34\scriptsize{±0.23} & 20.77\scriptsize{±1.31}& 22.19\scriptsize{±0.54} & 29.58\scriptsize{±0.15} & 24.41\scriptsize{±0.95}& \textbf{32.07\scriptsize{±0.27}}\\
     \hline

    \multirow{3}{*}{Domain-Incremental}& Digit10 ($\alpha = 0.1$) & 77.59\scriptsize{±0.39} & 79.09\scriptsize{±0.58} & 71.26\scriptsize{±0.04} & 76.44\scriptsize{±1.05} & 74.77\scriptsize{±0.23} & 77.59\scriptsize{±0.39} & 83.85\scriptsize{±0.80} & \textbf{85.96\scriptsize{±0.14}}\\
    & Office31 ($\alpha = 1$) & 39.25\scriptsize{±1.61} & 43.01\scriptsize{±1.59} & 37.44\scriptsize{±0.72} & 45.21\scriptsize{±2.10} & 37.55\scriptsize{±0.69} & 39.25\scriptsize{±1.61} & 46.26\scriptsize{±2.24} & \textbf{50.80\scriptsize{±0.77}}\\
    & DomainNet ($\alpha = 10$) & 51.73\scriptsize{±2.32} & 49.12\scriptsize{±2.71} & 46.30\scriptsize{±1.42} & 50.01\scriptsize{±3.31} & 41.76\scriptsize{±1.26} & 51.73\scriptsize{±2.32} & 47.28\scriptsize{±3.01} &\textbf{56.66\scriptsize{±0.50}} \\
    \bottomrule[1pt]
    \end{tabular}
    }
    \label{incremental}
    \vspace{-0.20cm}
\end{table*}

\vspace{2pt}
\noindent \textbf{Personalized Informative Model.} In FIL with Non-IID client samples, sample importance should be defined based not only on their importance in the local dataset but also on their correlation to the global dataset across clients such that the model can be better trained with the cached samples. In a standard FL scenario, each client can only access its own local model and global model, which respectively contains local and global information. A straightforward idea here is to calculate two sample importance scores using local and global models {and cache samples based on such two scores for data reply}. Then, one can build upon such an idea by adding following capabilities: (1) {The global model is aggregated by local models from participated clients and the gradient norm of the global model can be calculated locally without training the global model.} (2) A control mechanism should be available to adjust the proportion of local and global information in the sample importance.

Toward the above goals, here we introduce an additional personalized informative model (PIM) for each client, which digests both the information from the local model and the global model. Then, we propose to adopt a ratio factor to adjust the information proportion from local and global sides. Finally, we can record the gradient norms of the samples to calculate the importance scores during the update of PIM, resulting in sample importance scores with both local and global information. Suppose that the client $k$ receives the global model $w^{t-1}$ and then the $t$-th 
new task arrives, and the clients update PIM $v_k^{t-1}$ with previous local samples $\mathcal{T}^{t-1}_{k,local}$ in $s$ iterations as follows:
\begin{align}\label{personalizedmodel}
    v_{k,s}^{t-1} &= v_{k,s-1}^{t-1}-\eta\biggl(\sum_{i=1}^{M}\nabla l\left(f_{v_{k,s-1}^{t-1}}(\Tilde{x}_{k,t-1}^{(i)}),\Tilde{y}_{k,t-1}^{(i)}\right)\nonumber\\
    &+ q(\lambda)(v_{k,s-1}^{t-1}-w^{t-1})\biggl).
\end{align}
where $\ q(\lambda) =  \frac{1-\lambda}{2 \lambda},  \lambda \in (0,1)$,
 and $\eta$ is the rate to control the step size of the update. The hyper-parameter $\lambda$ adjusts the balance between the local and global information incorporated in the update.

To better understand the update, we can draw an analogy to momentum methods in optimization.
Momentum-based methods leverage the past updates to guide the current update direction \cite{chan1996momentum}. Similarly, the term $q(\lambda)(v_{k,s-1}^{t-1}-w^{t-1})$ acts as a momentum component. It incorporates information from the global model $w^{t-1}$ to influence the update of PIM $v_k^{t-1}$. The hyper-parameter $\lambda$ controls the weight of this momentum component, and it lies within the range of (0,1). When $\lambda$ is close to 0, PIM primarily focuses on recovering the global model $w^{t-1}$. In other words, it will align itself with the global data. On the other hand, as $\lambda$ becomes larger, it leads to a stronger emphasis on local training.

\begin{theorem}[Convergence of PIM]\label{thrm:2}
Assuming that the global model $w^t$ converges to the optimal model $\hat{w}$ at communication round $t$  by $g(t)$ as: $ \mathbb{E}\Big[||w^t-\hat{w}||^2\Big] \leq g(t)$, $\lim_{t\rightarrow\infty}g(t)=0$ and $g(t+1)\le g(t)$, there exists a constant $C < \infty$ such that for any client $k \in [K]$ the personalized informative model $v_k^t$ can converge to the optimal model $\hat{v_k}$ by $Cg(t)$.
\end{theorem}
With Theorem \ref{thrm:2}, we ensure the convergence of PIM and thus can calculate the gradient norms of samples during the training stage. We provide the proof in Appendix \ref{sec:thm2}.

\vspace{3pt}
\noindent \textbf{Sample Importance.} To quantify the sample importance, we investigate the impact of samples on the generalization ability of the model, and allocate the samples that can enhance the generalization with higher importance. We calculate the gradient norm with respect to model parameters of PIM as the importance scores to samples. 
The gradient norm of samples during training is recorded, which can be regarded as the contribution of the sample to the model update. A similar flavor of such a method can be found in \cite{DBLP:journals/corr/abs-1812-05159,DBLP:journals/corr/abs-2107-07075}.
Suppose that the client $k$ has converged on ${(t-1)}$-th tasks with local samples $\mathcal{T}^{t-1}_{k,local}$, when it comes to the incremental $t$-th task, the client should calculate the importance scores for all local samples. Here we denote $G^p(\Tilde{x}^{(i)}_{k,t-1})$ is the gradient norm of the sample $(\Tilde{x}^{(i)}_{k,t-1},\Tilde{y}^{(i)}_{k,t-1})$ in $p$-th iteration during the update of PIM $v^{t-1}_{k,p}$, which is: 
\begin{align}\label{importanceScore1}
    G^p(\Tilde{x}^{(i)}_{k,t-1}) = \left|\left|\nabla l\left(f_{v^{t-1}_{k,p}}(\Tilde{x}^{(i)}_{k,t-1}),\Tilde{y}^{(i)}_{k,t-1}\right)\right|\right|^2.
\end{align}
According to \cite{DBLP:journals/corr/abs-1812-05159}, the difference in the gradient of the loss function with and without a sample $(\Tilde{x}^{(j)}_{k,t-1},\Tilde{y}^{(j)}_{k,t-1})$ is upper bounded and the bound is linearly dependent on the sample gradient norm defined in Eq.~\ref{importanceScore1}. Thus, caching samples based on sample gradient norms can least affect the gradient and best preserve the dynamics of training.

As PIM integrates both local and global models, a greater gradient norm of a sample with PIM indicates that such a sample drives PIM more to fitting the task with local and global knowledge. Such an effect could be more prominent at early training during $s$ iterations for PIM, where fluctuation around optima is rare than that later in the training. Thus, we accumulate the gradient norm during the training of PIM and emphasis on the early training stage to calculate the sample importance as:
\begin{align}\label{importanceScore2}
I(\Tilde{x}^{(i)}_{k,t-1}) &= \sum_{p=1}^s\frac{1}{p}G^p(\Tilde{x}^{(i)}_{k,t-1}).
\end{align} 
We also provide illustrative experimental results with $I(\Tilde{x}^{(i)}_{k,t-1}) = \sum_{p=1}^s{G^p(\Tilde{x}^{(i)}_{k,t-1})}$ in the Appendix \ref{sec:result} and show the performance improvement with the sample importance adopted in Eq. ~\ref{importanceScore2}.

\vspace{3pt}
\noindent \textbf{Local Training.} After caching important samples with higher importance scores, each client continues to train the local model $w^t_k$ with local samples $\mathcal{T}^t_{k,local}$ in iteration $p\in[1,s]$ as follows:
\begin{align}\label{LocalTraining}
        w_{k,p}^{t} &= w_{k,p-1}^{t}-\eta\sum_{i=1}^{M}\nabla l\left(f_{w_{k,p-1}^{t}}(\Tilde{x}_{k,t}^{(i)}),\Tilde{y}_{k,t}^{(i)}\right).
\end{align}

\begin{figure*}[h]
  \centering
    \includegraphics[width=\textwidth]{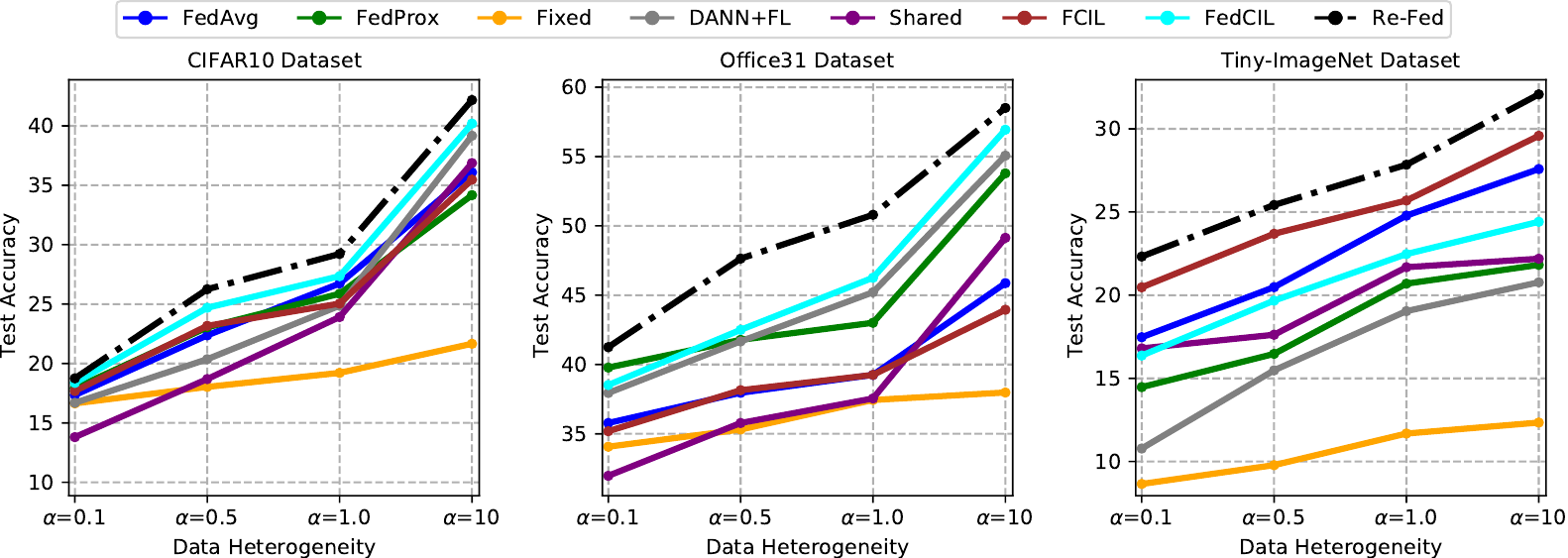}
  \caption{Performance w.r.t data heterogeneity $\alpha$ for three datasets. }
  \label{data_heterogeneity}
\end{figure*}
\begin{table*}[htbp]
    \centering
        \caption{Test accuracy for Re-Fed w.r.t data heterogeneity $\alpha$ and hyper-parameter $\lambda$ on CIFAR10, Office31 and Tiny-ImageNet.}
    \renewcommand\arraystretch{1.2}
      \resizebox{\linewidth}{!}{
    \begin{tabular}{c |c c c | c c c | c c c | c c c}
    \toprule[1pt]
         \multirow{2}{*}{Dataset}  & \multicolumn{3}{c|}{$\alpha=0.1$} & \multicolumn{3}{c|}{$\alpha=0.5$} & \multicolumn{3}{c|}{$\alpha=1.
         0$} & \multicolumn{3}{c}{$\alpha=10$} \\
       \cline{2-13}
        & $\lambda=0.2$ & $\lambda=0.5$ & $\lambda=0.8$ & $\lambda=0.2$ & $\lambda=0.5$ & $\lambda=0.8$ & $\lambda=0.2$ & $\lambda=0.5$ & $\lambda=0.8$ & $\lambda=0.2$ & $\lambda=0.5$ & $\lambda=0.8$ \\
       \hline
       CIFAR10& \textbf{18.75\scriptsize{±1.30}} & 18.61\scriptsize{±1.09} & 17.91\scriptsize{±0.81} & \textbf{26.25\scriptsize{±1.64}} & 26.00\scriptsize{±0.97} & 25.62\scriptsize{±0.52} & 27.05\scriptsize{±0.88} & 27.80\scriptsize{±0.21} & \textbf{29.22\scriptsize{±0.49}} & 38.43\scriptsize{±0.43} & 40.04\scriptsize{±0.19} & \textbf{42.17\scriptsize{±0.25}}  \\
       Office31& \textbf{41.25\scriptsize{±1.01}} & 39.29\scriptsize{±1.34} & 38.18\scriptsize{±0.68} & 46.86\scriptsize{±0.91} & \textbf{47.64\scriptsize{±0.53}} & 47.13\scriptsize{±1.16} & 43.81\scriptsize{±0.73} & 48.67\scriptsize{±0.99} & \textbf{50.08\scriptsize{±0.77}} & 52.79\scriptsize{±1.28} & 55.92\scriptsize{±0.38} & \textbf{58.51\scriptsize{±0.46}}\\
       Tiny-ImageNet& \textbf{22.32\scriptsize{±0.12}} & 20.51\scriptsize{±0.98} & 18.00\scriptsize{±1.30} & 24.60\scriptsize{±0.48} & \textbf{25.42\scriptsize{±0.59}} & 24.39\scriptsize{±0.66} & 24.88\scriptsize{±0.87} & 27.15\scriptsize{±0.78} & \textbf{27.84\scriptsize{±0.73}} & 29.03\scriptsize{±0.30} & 30.26\scriptsize{±0.24} & \textbf{32.07\scriptsize{±0.27}}\\
    \bottomrule[1pt]
    \end{tabular}}
    \label{lambda}
    \vspace{-0.2cm}
\end{table*}

\vspace{3pt}
\noindent \textbf{Modularity of Re-Fed.} From the Re-Fed framework and Algorithm \ref{simplified}, we can see that a key feature of Re-Fed is its unique modularity. One can readily use prior art developed for FIL algorithm, and employ Re-Fed as a useful off-the-shelf add-on. Our method has several advantages:
\begin{itemize}
    \item \textbf{Optimization:} It is possible to plug in other aggregation methods beyond FedAvg \cite{mcmahan2017communication} in Algorithm \ref{simplified} to update the global model, and inherit the convergence benefits. In the subsequent experimental design, we investigate the performance of our Re-Fed framework with FedAvg algorithm and provide a detailed algorithm definition using FedAvg in Appendix \ref{sec:detailed}.
    
    \item \textbf{Privacy:} Re-Fed transmits no more extra information over the network than typical FL algorithms. This is different from most other FIL methods where sample reconstruction methods are applied for data replay, which may raise privacy concerns.

    \item \textbf{Resource:} Re-Fed allows each client to train a backbone model using only its local training data, without employing additional distillation data or generated data, leading to extra either computation cost or storage overhead.
\end{itemize}

\begin{figure*}[h]
  \centering
    \includegraphics[width=\textwidth]{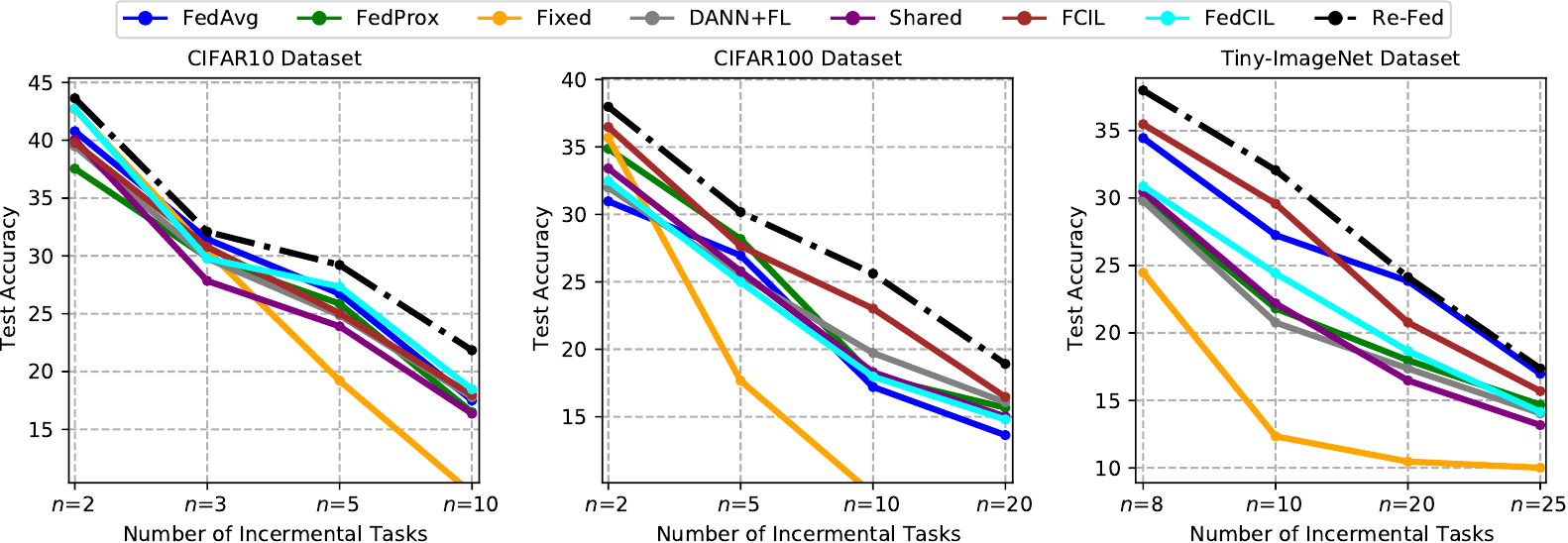}
  \caption{Performance w.r.t number of incremental tasks $n$ for three class-incremental datasets. }
  \label{ablation}
\end{figure*}
\begin{table*}[t]
    \renewcommand\arraystretch{1.25}
    \centering
    \caption{Evaluation of various methods, in terms of the communication rounds to reach the best test accuracy (150 communication rounds each task). We report the sum of communication rounds required to achieve the best performance on each task.}
      \resizebox{\linewidth}{!}{
    \begin{tabular}{c| c| c c |c c c |c c |c}
     \toprule[1pt]

     Scenario & Dataset & FedAvg & FedProx & Fixed & DANN+FL & Shared & FCIL & FedCIL & Re-Fed\\ \hline
    \multirow{3}{*}{Class-Incremental}& CIFAR10 (Task:5) & 613\scriptsize{±2.67} & 685\scriptsize{±3.00} & 142\scriptsize{±0.67} & 712\scriptsize{±3.67} & 574\scriptsize{±1.33} & 590\scriptsize{±2.67} & 738\scriptsize{±4.00} &\textbf{562\scriptsize{±1.67}}\\
    & CIFAR100 (Task:10) & 1103\scriptsize{±2.33} & 1246\scriptsize{±3.00} & 137\scriptsize{±2.00} & 1258\scriptsize{±4.67} & 1154\scriptsize{±3.33} & 1095\scriptsize{±2.67} & 1311\scriptsize{±5.67} &\textbf{1039\scriptsize{±4.33}}\\
     & Tiny-ImageNet (Task:10) & 1197\scriptsize{±2.67} & 1234\scriptsize{±2.67} & 132\scriptsize{±3.00} & 1305\scriptsize{±3.67} & 1278\scriptsize{±4.33} & 1185\scriptsize{±2.33} & 1317\scriptsize{±3.33} & \textbf{1128\scriptsize{±3.67}}\\

     \hline

    \multirow{3}{*}{Domain-Incremental}& Digit10 (Task:4) & 410\scriptsize{±1.67} & 412\scriptsize{±0.67} & 112\scriptsize{±0.33} & 483\scriptsize{±1.33} & 372\scriptsize{±2.00} & 410\scriptsize{±1.67} & 419\scriptsize{±2.67} & \textbf{325\scriptsize{±1.33}}\\
    & Office31 (Task:3) & 413\scriptsize{±2.67} & 429\scriptsize{±2.00} & 144\scriptsize{±0.67} & 436\scriptsize{±3.67} & 391\scriptsize{±1.12} & 413\scriptsize{±2.67} & 431\scriptsize{±3.33} & \textbf{388\scriptsize{±1.67}}\\
    & DomainNet (Task:6) & 726\scriptsize{±3.33} & 767\scriptsize{±2.67} & 141\scriptsize{±1.67} & 752\scriptsize{±4.00} & 694\scriptsize{±2.67} & 726\scriptsize{±3.33} & 791\scriptsize{±3.67} & \textbf{661\scriptsize{±2.33}}\\
    \bottomrule[1pt]
    \end{tabular}
    }
    \label{communication}
\vspace{-0.2cm}
\end{table*}

\begin{figure*}[h]
  \centering
  \includegraphics[width=\textwidth]{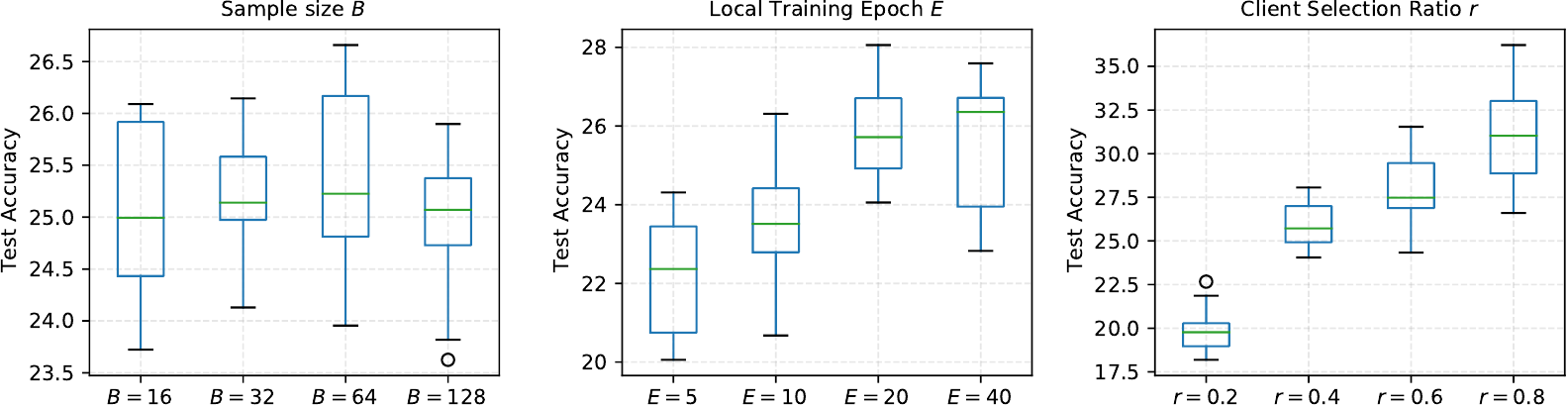}
  \caption{Performance of Re-Fed under different configurations (a) local training epoch $E$, (b) sample size $B$ in the classifier, (c) client selection ratio $r$ of all clients on CIFAR100 with $\alpha$ = 5.0.}
  \label{parameter}
\end{figure*}
\begin{table*}[h]
    \centering
        \caption{Test accuracy for Re-Fed and other baselines w.r.t memory size $M$ on Digit10, CIFAR10, CIFAR100 and Tiny-ImageNet.}
    \renewcommand\arraystretch{1.2}
      \resizebox{\linewidth}{!}{
    \begin{tabular}{c |c c c | c c c | c c c | c c c}
    \toprule[1pt]
         \multirow{2}{*}{Dataset} & \multicolumn{3}{c|}{FedAvg} & \multicolumn{3}{c|}{DANN+FL}&\multicolumn{3}{c|}{FCIL} & \multicolumn{3}{c}{Re-Fed} \\
        \cline{2-13}
        & $M=1000$ & $M=2000$ & $M=3000$ & $M=1000$ & $M=2000$ & $M=3000$ & $M=1000$ & $M=2000$ & $M=3000$ & $M=1000$ & $M=2000$ & $M=3000$\\
       \hline
       Digit10& 77.05\scriptsize{±0.24} & 77.59\scriptsize{±0.39} & 81.64\scriptsize{±0.31} & 75.16\scriptsize{±0.86} & 76.44\scriptsize{±1.05} & 83.34\scriptsize{±0.97} & 77.50\scriptsize{±0.24} & 77.59\scriptsize{±0.39} & 81.64\scriptsize{±0.31} & \textbf{82.30\scriptsize{±0.16}} & \textbf{85.96\scriptsize{±0.14}} & \textbf{86.32\scriptsize{±0.04}}\\
       CIFAR10& 26.73\scriptsize{±1.12} & 30.19\scriptsize{±1.63} & 33.41\scriptsize{±0.92} & 24.86\scriptsize{±2.31} & 27.65\scriptsize{±1.34} & 30.09\scriptsize{±1.06} & 25.04\scriptsize{±0.11} & 29.14\scriptsize{±0.19} & 31.77\scriptsize{±0.28} & \textbf{29.22\scriptsize{±0.49}} & \textbf{32.83\scriptsize{±0.20}} & \textbf{33.41\scriptsize{±0.74}}  \\
       CIFAR100& 17.21\scriptsize{±1.35} & 23.58\scriptsize{±1.82} & 29.90\scriptsize{±2.11} & 19.73\scriptsize{±2.17} & 25.56\scriptsize{±2.68} & 28.49\scriptsize{±1.98} & 23.02\scriptsize{±0.66} & 26.12\scriptsize{±0.54} & 29.06\scriptsize{±0.89} & \textbf{25.61\scriptsize{±0.88}} & \textbf{28.41\scriptsize{±0.24}} & \textbf{29.90\scriptsize{±0.71}} \\
       Tiny-ImageNet& 24.42\scriptsize{±0.59} & 27.58\scriptsize{±0.74} & 31.50\scriptsize{±0.63} & 18.06\scriptsize{±1.08} & 20.77\scriptsize{±1.31} & 26.93\scriptsize{±0.77} & 25.20\scriptsize{±1.12} & 29.58\scriptsize{±0.15} & 33.44\scriptsize{±0.38} & \textbf{28.31\scriptsize{±0.19}} & \textbf{32.07\scriptsize{±0.27}} & \textbf{35.66\scriptsize{±0.42}}\\
    \bottomrule[1pt]
    \end{tabular}}
    \label{memory}
\vspace{-0.25cm}
\end{table*}

\section{Experiments}
In this section, we evaluate our proposed framework concerning two incremental learning scenarios. We investigate the relationship between the data heterogeneity and the balance between the local and global information incorporated in PIM. Additionally, we conduct parameter sensitivity analysis to verify the effectiveness of our method. 

\subsection{Experiment Setup}
\noindent\textbf{Dataset:} We conduct our experiments with heterogeneously partitioned datasets over two federated incremental scenarios on six datasets. (1) \textbf{Class-Incremental Learning:} CIFAR10\cite{krizhevsky2009learning}, CIFAR100\cite{krizhevsky2009learning} and Tiny-ImageNet\cite{le2015tiny}. (2) \textbf{Domain-Incremental Learning:} Digit10, Office31\cite{saenko2010adapting} and DomainNet\cite{peng2019moment}. Among them, the Digit10 dataset contains 10 digit categories in four domains: MNIST\cite{lecun2010mnist}, EMNIST\cite{cohen2017emnist}, USPS\cite{hull1994database} and SVHN\cite{netzer2011reading}. Details of datasets and data processing can be found in Appendix \ref{sec:dataset}. 

\noindent\textbf{Baseline:} For a fair comparison with other key works, we follow the same protocols proposed by \cite{mcmahan2017communication,rebuffi2017icarl} to set up FIL tasks. We evaluate all the methods with two representative FL models \textbf{FedAvg} \cite{mcmahan2017communication} and \textbf{Fedprox} \cite{li2020federated}, two models designed for federated class-incremental learning: \textbf{FCIL} \cite{dong2022federated} and \textbf{FedCIL} \cite{qi2023better}, and three customed methods of \textbf{Fixed}: we train the model only from the first task and evaluate it for all the coming sequence of tasks; \textbf{DANN+FL}: here we adopt the robust adversarial-based method DANN\cite{ganin2016domain} in local training for domain adaption; \textbf{Shared}: we adopt all front layers before the last fully connected layer as shared layers, and use relevant different fully-connected layers to obtain outputs for different tasks. Details of baselines can be found in Appendix \ref{sec:baseline}.

\noindent \textbf{Configurations:}
Unless otherwise mentioned, we set the number of local training epoch ${E} = 20$ and communication round ${T} = 150$ for each task, which ensures the convergence of previous tasks before the arrival of new task. We use the Dirichlet distribution Dir($\alpha$) to distribute local samples to yield data heterogeneity for all tasks {where a smaller $\alpha$ indicates higher data heterogeneity}. We employ ResNet18 \cite{he2016deep} as the basic backbone model in all methods. We calculate the Top-10 accuracy for the Tiny-ImageNet and DomainNet datasets and Top-1 accuracy for others. Each experiment setting is run three times and we record accuracy in the final 10 rounds and report the average value and standard deviation. The total clients number is 20/10 with an active ratio $\emph{k} = 0.4$ for \{CIFAR10, CIFAR100, Tiny-ImageNet, Digit10, DomainNet\}/\{Office31\}. The maximum number of cached samples $M$ is 2000/1000/300 for \{Digit10, DomainNet, Tiny-ImageNet\}/\{CIFAR10, CIFAR100\}/\{Office31\}. We experiment with different numbers of incremental tasks and each task arrives with new classes: 10/10/5 tasks with 20/10/2 new classes in each task for \{Tiny-ImageNet\}/ \{CIFAR100\}/ \{CIFAR10\}. Details of configurations can be found in Appendix \ref{sec:config}.

\subsection{Performance Overview}
\noindent \textbf{Test Accuracy.}
Table \ref{incremental} shows the test accuracy of various methods with data heterogeneity across six datasets. We report the final accuracy of the global model when all clients finish their training on all tasks. FedCIL outperforms FedAvg and FedProx on CIFAR10, Digit10 and Office31 as the generator in FedCIL exhibits effective training on simple datasets, enabling the generation of high-quality samples for data replay. However, its performance experiences a significant decline with larger-scale datasets such as CIFAR100 and DomainNet, where the classes and domains become more complex. In the federated domain-incremental learning scenario, FCIL reverts to the FedAvg algorithm when there are no new incremental sample classes. Re-Fed achieves the best performance in all cases by a margin of 1.87\%$\sim$19.73\% in terms of final accuracy. 
More discussions and results on model performance are available in Appendix \ref{sec:result}.

\noindent \textbf{Data Heterogeneity.}
Figure \ref{data_heterogeneity} displays the test accuracy with different levels of data heterogeneity on three datasets. As shown in this figure, all methods achieve an improvement in test accuracy with the decline in data heterogeneity, and Re-Fed consistently achieves a leading improvement in performance with different levels of data heterogeneity.

Then, we conduct more research on the setting of hyper-parameter $\lambda$. In our framework, we modify the $\lambda$ value to adjust the global and local information proportion in PIM. As shown in Table \ref{lambda}, we select three different $\lambda$ values with four different data heterogeneity settings and evaluate the final test accuracy on three datasets. Experimental results show that the value of $\lambda$ should be chosen accordingly under different data heterogeneity. Nevertheless, results exhibit the same trend: as the degree of data heterogeneity increases, our Re-Fed framework performs better while $\lambda$ decreases as PIM contains more global information. Empirically, striking a balance between global and local information is the key to addressing the data heterogeneity in FIL. Vice versa, as $\alpha$ increases, the data distribution on the client side becomes more IID. At this point, the clients require less global information and can rely more on their local information for caching important samples. 

\noindent \textbf{Quantitative Analysis}.
Figure \ref{ablation} shows the qualitative analysis of the number of incremental tasks $n$ on three class-incremental datasets. According to these curves, we can easily observe that our model performs better than other baselines across all tasks, with varying numbers of incremental tasks. It demonstrates that Re-Fed enables clients to learn new incremental classes better than other methods.

\noindent \textbf{Communication Efficiency.}
Table \ref{communication} shows the evaluation of various methods in terms of the communication rounds to reach the {test accuracy reported in Table \ref{incremental}}. Here we show the sum of communication rounds required to achieve the performance on each task. Re-Fed {requires the least communication rounds to achieve the reported test accuracy} on all datasets with its caching method. Thus, the local model is easier to converge. More details of the communication round on each task are available in Appendix \ref{sec:result}.

\noindent \textbf{Parameter Sensitivity Analysis.}
Figure \ref{parameter} shows the performance of Re-Fed under different configurations with standard boxplots from ten trails with different random seeds. Re-Fed achieves similar performance with different sample sizes $B$, and it achieves a better result when we increase the local training epochs. However, Re-Fed has a comparable performance when the $E$ is set to 20 and 40. In addition, a larger client selection ratio $r$ contributes to higher test accuracy.

For the FIL scenario, we conduct additional research on the client storage $M$. When $M$ is large enough for clients to cache samples from all tasks, our framework degrades to a normal FedAvg algorithm with a naive caching method. As shown in Table \ref{memory}, we select three different $M$ values to conduct experiments. Experimental results demonstrate that larger memory size $M$ contributes to the training and Re-Fed outperforms other baselines in all cases. To conclude, Re-Fed is only sensitive to few parameters and still robust to most parameters in a large range.

\section{Conclusion and Future Work}
We proposed Re-Fed, a simple framework, to address the catastrophic forgetting with data heterogeneity in federated incremental learning. Re-Fed can be thought of as a lightweight personalization add-on for any federated learning algorithms with global aggregation, which maintains privacy and communication efficiency. Extensive experiments conducted on various settings and baselines show that Re-Fed achieves significant improvement in test accuracy. 

Although existing works and our method have demonstrated great effectiveness over the FIL scenarios, none have studied the dynamic requirements of the edge clients. To deploy the FL system in practical settings, it is necessary to consider personalized local factors such as storage, computation and even the different arrival time of the new task. In the future, we seek to work a step forward in this field.

\section*{Acknowledgements}
This work is supported by National Natural Science Foundation of China under grants 62376103, 62302184, 62206102, Science and Technology Support Program of Hubei Province under grant 2022BAA046, and CCF-AFSG Research Fund.

{
    \small
    \bibliographystyle{ieeenat_fullname}
    \bibliography{main}

\begin{thebibliography}{51}
\providecommand{\natexlab}[1]{#1}
\providecommand{\url}[1]{\texttt{#1}}
\expandafter\ifx\csname urlstyle\endcsname\relax
  \providecommand{\doi}[1]{doi: #1}\else
  \providecommand{\doi}{doi: \begingroup \urlstyle{rm}\Url}\fi

\bibitem[Chan et~al.(1996)Chan, Jegadeesh, and Lakonishok]{chan1996momentum}
Louis~KC Chan, Narasimhan Jegadeesh, and Josef Lakonishok.
\newblock Momentum strategies.
\newblock \emph{The journal of Finance}, 51\penalty0 (5):\penalty0 1681--1713, 1996.

\bibitem[Churamani et~al.(2021)Churamani, Kara, and Gunes]{Churamani2021DomainIncrementalCL}
Nikhil Churamani, Ozgur Kara, and Hatice Gunes.
\newblock Domain-incremental continual learning for mitigating bias in facial expression and action unit recognition.
\newblock \emph{ArXiv}, abs/2103.08637, 2021.

\bibitem[Cohen et~al.(2017)Cohen, Afshar, Tapson, and Van~Schaik]{cohen2017emnist}
Gregory Cohen, Saeed Afshar, Jonathan Tapson, and Andre Van~Schaik.
\newblock Emnist: Extending mnist to handwritten letters.
\newblock In \emph{2017 international joint conference on neural networks (IJCNN)}, pages 2921--2926. IEEE, 2017.

\bibitem[Criado et~al.(2022)Criado, Casado, Iglesias, Regueiro, and Barro]{criado2022non}
Marcos~F Criado, Fernando~E Casado, Roberto Iglesias, Carlos~V Regueiro, and Sen{\'e}n Barro.
\newblock Non-iid data and continual learning processes in federated learning: A long road ahead.
\newblock \emph{Information Fusion}, 88:\penalty0 263--280, 2022.

\bibitem[Dantam et~al.(2016)Dantam, Kingston, Chaudhuri, and Kavraki]{Dantam2016IncrementalTA}
Neil~T. Dantam, Zachary~K. Kingston, Swarat Chaudhuri, and Lydia~E. Kavraki.
\newblock Incremental task and motion planning: A constraint-based approach.
\newblock In \emph{Robotics: Science and Systems}, 2016.

\bibitem[Dong et~al.(2022)Dong, Wang, Fang, Sun, Xu, Wang, and Zhu]{dong2022federated}
Jiahua Dong, Lixu Wang, Zhen Fang, Gan Sun, Shichao Xu, Xiao Wang, and Qi Zhu.
\newblock Federated class-incremental learning.
\newblock In \emph{Proceedings of the IEEE/CVF Conference on Computer Vision and Pattern Recognition}, pages 10164--10173, 2022.

\bibitem[Du et~al.(2020)Du, Wu, Yoshinaga, Yau, Ji, and Li]{du2020federated}
Zhaoyang Du, Celimuge Wu, Tsutomu Yoshinaga, Kok-Lim~Alvin Yau, Yusheng Ji, and Jie Li.
\newblock Federated learning for vehicular internet of things: Recent advances and open issues.
\newblock \emph{IEEE Open Journal of the Computer Society}, 1:\penalty0 45--61, 2020.

\bibitem[Fernando et~al.(2017)Fernando, Banarse, Blundell, Zwols, Ha, Rusu, Pritzel, and Wierstra]{Fernando2017PathNetEC}
Chrisantha Fernando, Dylan~S. Banarse, Charles Blundell, Yori Zwols, David~R Ha, Andrei~A. Rusu, Alexander Pritzel, and Daan Wierstra.
\newblock Pathnet: Evolution channels gradient descent in super neural networks.
\newblock \emph{ArXiv}, abs/1701.08734, 2017.

\bibitem[Ganin et~al.(2016)Ganin, Ustinova, Ajakan, Germain, Larochelle, Laviolette, Marchand, and Lempitsky]{ganin2016domain}
Yaroslav Ganin, Evgeniya Ustinova, Hana Ajakan, Pascal Germain, Hugo Larochelle, Fran{\c{c}}ois Laviolette, Mario Marchand, and Victor Lempitsky.
\newblock Domain-adversarial training of neural networks.
\newblock \emph{The journal of machine learning research}, 17\penalty0 (1):\penalty0 2096--2030, 2016.

\bibitem[Hanzely and Richt{\'{a}}rik(2020)]{DBLP:journals/corr/abs-2002-05516}
Filip Hanzely and Peter Richt{\'{a}}rik.
\newblock Federated learning of a mixture of global and local models.
\newblock \emph{CoRR}, abs/2002.05516, 2020.

\bibitem[He et~al.(2016)He, Zhang, Ren, and Sun]{he2016deep}
Kaiming He, Xiangyu Zhang, Shaoqing Ren, and Jian Sun.
\newblock Deep residual learning for image recognition.
\newblock In \emph{Proceedings of the IEEE conference on computer vision and pattern recognition}, pages 770--778, 2016.

\bibitem[Hsu et~al.(2018)Hsu, Liu, and Kira]{Hsu2018ReevaluatingCL}
Yen-Chang Hsu, Yen-Cheng Liu, and Zsolt Kira.
\newblock Re-evaluating continual learning scenarios: A categorization and case for strong baselines.
\newblock \emph{ArXiv}, abs/1810.12488, 2018.

\bibitem[Hull(1994)]{hull1994database}
Jonathan~J. Hull.
\newblock A database for handwritten text recognition research.
\newblock \emph{IEEE Transactions on pattern analysis and machine intelligence}, 16\penalty0 (5):\penalty0 550--554, 1994.

\bibitem[Jeong et~al.(2018{\natexlab{a}})Jeong, Oh, Kim, Park, Bennis, and Kim]{DBLP:journals/corr/abs-1811-11479}
Eunjeong Jeong, Seungeun Oh, Hyesung Kim, Jihong Park, Mehdi Bennis, and Seong{-}Lyun Kim.
\newblock Communication-efficient on-device machine learning: Federated distillation and augmentation under non-iid private data.
\newblock \emph{CoRR}, abs/1811.11479, 2018{\natexlab{a}}.

\bibitem[Jeong et~al.(2018{\natexlab{b}})Jeong, Oh, Kim, Park, Bennis, and Kim]{Jeong2018CommunicationEfficientOM}
Eunjeong Jeong, Seungeun Oh, Hyesung Kim, Jihong Park, Mehdi Bennis, and Seong-Lyun Kim.
\newblock Communication-efficient on-device machine learning: Federated distillation and augmentation under non-iid private data.
\newblock \emph{ArXiv}, abs/1811.11479, 2018{\natexlab{b}}.

\bibitem[Jung et~al.(2020)Jung, Ahn, Cha, and Moon]{Jung2020ContinualLW}
Sangwon Jung, Hongjoon Ahn, Sungmin Cha, and Taesup Moon.
\newblock Continual learning with node-importance based adaptive group sparse regularization.
\newblock \emph{arXiv: Learning}, 2020.

\bibitem[Krizhevsky et~al.(2009)Krizhevsky, Hinton, et~al.]{krizhevsky2009learning}
Alex Krizhevsky, Geoffrey Hinton, et~al.
\newblock Learning multiple layers of features from tiny images.
\newblock 2009.

\bibitem[Le and Yang(2015)]{le2015tiny}
Ya Le and Xuan Yang.
\newblock Tiny imagenet visual recognition challenge.
\newblock \emph{CS 231N}, 7\penalty0 (7):\penalty0 3, 2015.

\bibitem[LeCun et~al.(2010)LeCun, Cortes, and Burges]{lecun2010mnist}
Yann LeCun, Corinna Cortes, and Chris Burges.
\newblock Mnist handwritten digit database, 2010.

\bibitem[Lei et~al.(2022)Lei, Wang, Zhong, Wang, and Ng]{lei2022federated}
Yuan Lei, Shir~Li Wang, Minghui Zhong, Meixia Wang, and Theam~Foo Ng.
\newblock A federated learning framework based on incremental weighting and diversity selection for internet of vehicles.
\newblock \emph{Electronics}, 11\penalty0 (22):\penalty0 3668, 2022.

\bibitem[Li et~al.(2020{\natexlab{a}})Li, Hu, Beirami, and Smith]{DBLP:journals/corr/abs-2012-04221}
Tian Li, Shengyuan Hu, Ahmad Beirami, and Virginia Smith.
\newblock Federated multi-task learning for competing constraints.
\newblock \emph{CoRR}, abs/2012.04221, 2020{\natexlab{a}}.

\bibitem[Li et~al.(2020{\natexlab{b}})Li, Sahu, Zaheer, Sanjabi, Talwalkar, and Smith]{li2020federated}
Tian Li, Anit~Kumar Sahu, Manzil Zaheer, Maziar Sanjabi, Ameet Talwalkar, and Virginia Smith.
\newblock Federated optimization in heterogeneous networks.
\newblock \emph{Proceedings of Machine Learning and Systems}, 2:\penalty0 429--450, 2020{\natexlab{b}}.

\bibitem[Li et~al.()Li, Xu, Song, Li, Li, Shao, and Zhan]{DBLP:conf/cvpr/LiXSLLSZ22}
Xin{-}Chun Li, Yi{-}Chu Xu, Shaoming Song, Bingshuai Li, Yinchuan Li, Yunfeng Shao, and De{-}Chuan Zhan.
\newblock Federated learning with position-aware neurons.
\newblock In \emph{{IEEE/CVF} Conference on Computer Vision and Pattern Recognition, {CVPR} 2022, New Orleans, LA, USA, June 18-24, 2022}, pages 10072--10081.

\bibitem[Li et~al.(2021)Li, Tao, Zhang, Liu, and Xu]{li2021privacy}
Yijing Li, Xiaofeng Tao, Xuefei Zhang, Junjie Liu, and Jin Xu.
\newblock Privacy-preserved federated learning for autonomous driving.
\newblock \emph{IEEE Transactions on Intelligent Transportation Systems}, 23\penalty0 (7):\penalty0 8423--8434, 2021.

\bibitem[Lin et~al.(2020)Lin, Kong, Stich, and Jaggi]{lin2020ensemble}
Tao Lin, Lingjing Kong, Sebastian~U Stich, and Martin Jaggi.
\newblock Ensemble distillation for robust model fusion in federated learning.
\newblock \emph{Advances in Neural Information Processing Systems}, 33:\penalty0 2351--2363, 2020.

\bibitem[Liu et~al.(2019)Liu, Zhang, Song, and Letaief]{Liu2019edgeAssistedHF}
Lumin Liu, Jun Zhang, S.~H. Song, and Khaled~Ben Letaief.
\newblock Edge-assisted hierarchical federated learning with non-iid data.
\newblock \emph{ArXiv}, abs/1905.06641, 2019.

\bibitem[Liu et~al.(2020)Liu, Wu, Menta, Herranz, Raducanu, Bagdanov, Jui, and de~Weijer]{liu2020generative}
Xialei Liu, Chenshen Wu, Mikel Menta, Luis Herranz, Bogdan Raducanu, Andrew~D Bagdanov, Shangling Jui, and Joost~van de Weijer.
\newblock Generative feature replay for class-incremental learning.
\newblock In \emph{Proceedings of the IEEE/CVF Conference on Computer Vision and Pattern Recognition Workshops}, pages 226--227, 2020.

\bibitem[Long et~al.(2015)Long, Cao, Wang, and Jordan]{Long2015LearningTF}
Mingsheng Long, Yue Cao, Jianmin Wang, and Michael~I. Jordan.
\newblock Learning transferable features with deep adaptation networks.
\newblock \emph{ArXiv}, abs/1502.02791, 2015.

\bibitem[Ma et~al.(2022)Ma, Xie, Wang, Chen, and Shou]{ijcai2022p0303}
Yuhang Ma, Zhongle Xie, Jue Wang, Ke Chen, and Lidan Shou.
\newblock Continual federated learning based on knowledge distillation.
\newblock In \emph{Proceedings of the Thirty-First International Joint Conference on Artificial Intelligence, {IJCAI-22}}, pages 2182--2188. International Joint Conferences on Artificial Intelligence Organization, 2022.
\newblock Main Track.

\bibitem[Maltoni and Lomonaco(2018)]{Maltoni2018ContinuousLI}
Davide Maltoni and Vincenzo Lomonaco.
\newblock Continuous learning in single-incremental-task scenarios.
\newblock \emph{Neural networks : the official journal of the International Neural Network Society}, 116:\penalty0 56--73, 2018.

\bibitem[McMahan et~al.(2017)McMahan, Moore, Ramage, Hampson, and y~Arcas]{mcmahan2017communication}
Brendan McMahan, Eider Moore, Daniel Ramage, Seth Hampson, and Blaise~Aguera y Arcas.
\newblock Communication-efficient learning of deep networks from decentralized data.
\newblock In \emph{Artificial intelligence and statistics}, pages 1273--1282. PMLR, 2017.

\bibitem[Mensink et~al.(2013)Mensink, Verbeek, Perronnin, and Csurka]{mensink2013distance}
Thomas Mensink, Jakob Verbeek, Florent Perronnin, and Gabriela Csurka.
\newblock Distance-based image classification: Generalizing to new classes at near-zero cost.
\newblock \emph{IEEE transactions on pattern analysis and machine intelligence}, 35\penalty0 (11):\penalty0 2624--2637, 2013.

\bibitem[Mirza et~al.(2022)Mirza, Masana, Possegger, and Bischof]{mirza2022efficient}
M~Jehanzeb Mirza, Marc Masana, Horst Possegger, and Horst Bischof.
\newblock An efficient domain-incremental learning approach to drive in all weather conditions.
\newblock In \emph{Proceedings of the IEEE/CVF Conference on Computer Vision and Pattern Recognition}, pages 3001--3011, 2022.

\bibitem[Netzer et~al.(2011)Netzer, Wang, Coates, Bissacco, Wu, and Ng]{netzer2011reading}
Yuval Netzer, Tao Wang, Adam Coates, Alessandro Bissacco, Bo Wu, and Andrew~Y Ng.
\newblock Reading digits in natural images with unsupervised feature learning.
\newblock 2011.

\bibitem[Nguyen et~al.(2022)Nguyen, Pham, Pathirana, Ding, Seneviratne, Lin, Dobre, and Hwang]{nguyen2022federated}
Dinh~C Nguyen, Quoc-Viet Pham, Pubudu~N Pathirana, Ming Ding, Aruna Seneviratne, Zihuai Lin, Octavia Dobre, and Won-Joo Hwang.
\newblock Federated learning for smart healthcare: A survey.
\newblock \emph{ACM Computing Surveys (CSUR)}, 55\penalty0 (3):\penalty0 1--37, 2022.

\bibitem[Paul et~al.(2021)Paul, Ganguli, and Dziugaite]{DBLP:journals/corr/abs-2107-07075}
Mansheej Paul, Surya Ganguli, and Gintare~Karolina Dziugaite.
\newblock Deep learning on a data diet: Finding important examples early in training.
\newblock \emph{CoRR}, abs/2107.07075, 2021.

\bibitem[Peng et~al.(2019)Peng, Bai, Xia, Huang, Saenko, and Wang]{peng2019moment}
Xingchao Peng, Qinxun Bai, Xide Xia, Zijun Huang, Kate Saenko, and Bo Wang.
\newblock Moment matching for multi-source domain adaptation.
\newblock In \emph{Proceedings of the IEEE/CVF international conference on computer vision}, pages 1406--1415, 2019.

\bibitem[Qi et~al.(2023)Qi, Zhao, and Li]{qi2023better}
Daiqing Qi, Handong Zhao, and Sheng Li.
\newblock Better generative replay for continual federated learning.
\newblock \emph{arXiv preprint arXiv:2302.13001}, 2023.

\bibitem[Rebuffi et~al.(2017)Rebuffi, Kolesnikov, Sperl, and Lampert]{rebuffi2017icarl}
Sylvestre-Alvise Rebuffi, Alexander Kolesnikov, Georg Sperl, and Christoph~H Lampert.
\newblock icarl: Incremental classifier and representation learning.
\newblock In \emph{Proceedings of the IEEE conference on Computer Vision and Pattern Recognition}, pages 2001--2010, 2017.

\bibitem[Ruder(2017)]{ruder2017overview}
Sebastian Ruder.
\newblock An overview of multi-task learning in deep neural networks.
\newblock \emph{arXiv preprint arXiv:1706.05098}, 2017.

\bibitem[Saenko et~al.(2010)Saenko, Kulis, Fritz, and Darrell]{saenko2010adapting}
Kate Saenko, Brian Kulis, Mario Fritz, and Trevor Darrell.
\newblock Adapting visual category models to new domains.
\newblock In \emph{Computer Vision--ECCV 2010: 11th European Conference on Computer Vision, Heraklion, Crete, Greece, September 5-11, 2010, Proceedings, Part IV 11}, pages 213--226. Springer, 2010.

\bibitem[Toneva et~al.(2018)Toneva, Sordoni, des Combes, Trischler, Bengio, and Gordon]{DBLP:journals/corr/abs-1812-05159}
Mariya Toneva, Alessandro Sordoni, Remi~Tachet des Combes, Adam Trischler, Yoshua Bengio, and Geoffrey~J. Gordon.
\newblock An empirical study of example forgetting during deep neural network learning.
\newblock \emph{CoRR}, abs/1812.05159, 2018.

\bibitem[van~de Ven and Tolias(2019)]{vandeVen2019ThreeSF}
Gido~M. van~de Ven and Andreas~Savas Tolias.
\newblock Three scenarios for continual learning.
\newblock \emph{ArXiv}, abs/1904.07734, 2019.

\bibitem[Wang et~al.()Wang, Chen, Wang, and Wang]{DBLP:conf/cvpr/WangCW022}
Chunnan Wang, Xiang Chen, Junzhe Wang, and Hongzhi Wang.
\newblock {ATPFL:} automatic trajectory prediction model design under federated learning framework.
\newblock In \emph{{IEEE/CVF} Conference on Computer Vision and Pattern Recognition, {CVPR} 2022, New Orleans, LA, USA, June 18-24, 2022}, pages 6553--6562.

\bibitem[Wang et~al.(2023)Wang, Li, Xu, Li, Zhan, and Zeng]{wang2023dafkd}
Haozhao Wang, Yichen Li, Wenchao Xu, Ruixuan Li, Yufeng Zhan, and Zhigang Zeng.
\newblock Dafkd: Domain-aware federated knowledge distillation.
\newblock In \emph{Proceedings of the IEEE/CVF Conference on Computer Vision and Pattern Recognition}, pages 20412--20421, 2023.

\bibitem[Xu et~al.(2021)Xu, Glicksberg, Su, Walker, Bian, and Wang]{xu2021federated}
Jie Xu, Benjamin~S Glicksberg, Chang Su, Peter Walker, Jiang Bian, and Fei Wang.
\newblock Federated learning for healthcare informatics.
\newblock \emph{Journal of Healthcare Informatics Research}, 5:\penalty0 1--19, 2021.

\bibitem[Yin et~al.(2020)Yin, Farajtabar, and Li]{Yin2020SOLACL}
Dong Yin, Mehrdad Farajtabar, and Ang Li.
\newblock Sola: Continual learning with second-order loss approximation.
\newblock \emph{ArXiv}, abs/2006.10974, 2020.

\bibitem[Yoon et~al.(2021)Yoon, Jeong, Lee, Yang, and Hwang]{yoon2021federated}
Jaehong Yoon, Wonyong Jeong, Giwoong Lee, Eunho Yang, and Sung~Ju Hwang.
\newblock Federated continual learning with weighted inter-client transfer.
\newblock In \emph{International Conference on Machine Learning}, pages 12073--12086. PMLR, 2021.

\bibitem[Yu et~al.(2020)Yu, Twardowski, Liu, Herranz, Wang, Cheng, Jui, and van~de Weijer]{Yu2020SemanticDC}
Lu Yu, Bartlomiej Twardowski, Xialei Liu, Luis Herranz, Kai Wang, Yongmei Cheng, Shangling Jui, and Joost van~de Weijer.
\newblock Semantic drift compensation for class-incremental learning.
\newblock \emph{2020 IEEE/CVF Conference on Computer Vision and Pattern Recognition (CVPR)}, pages 6980--6989, 2020.

\bibitem[Zhang et~al.()Zhang, Shen, Ding, Tao, and Duan]{DBLP:conf/cvpr/ZhangS0TD22}
Lin Zhang, Li Shen, Liang Ding, Dacheng Tao, and Ling{-}Yu Duan.
\newblock Fine-tuning global model via data-free knowledge distillation for non-iid federated learning.
\newblock In \emph{{IEEE/CVF} Conference on Computer Vision and Pattern Recognition, {CVPR} 2022, New Orleans, LA, USA, June 18-24, 2022}, pages 10164--10173.

\bibitem[Zhu et~al.(2021)Zhu, Hong, and Zhou]{zhu2021data}
Zhuangdi Zhu, Junyuan Hong, and Jiayu Zhou.
\newblock Data-free knowledge distillation for heterogeneous federated learning.
\newblock In \emph{International Conference on Machine Learning}, pages 12878--12889. PMLR, 2021.

\end{thebibliography}
}
\appendix
\clearpage
\setcounter{page}{1}
\maketitlesupplementary

\section{Dataset}\label{sec:dataset}
\textbf{Class-Incremental Task Dataset:} New classes are incrementally introduced over time. The dataset starts with a subset of classes, and new classes are added in subsequent stages, allowing models to learn and adapt to an increasing number of classes.
\begin{itemize}
    \item \textbf{CIFAR10:} A dataset with 10 object classes, including various common objects, animals, and vehicles. It consists of 50,000 training images and 10,000 test images.

    \item \textbf{CIFAR100:} Similar to CIFAR10, but with 100 fine-grained object classes. It has 50,000 training images and 10,000 test images.

    \item \textbf{Tiny-ImageNet:} A subset of the ImageNet dataset with 200 object classes. It contains 100,000 training images, 10,000 validation images, and 10,000 test images.
\end{itemize}

\noindent \textbf{Domain-Incremental Task Dataset:} New domains are introduced gradually. The dataset initially contains samples from a specific domain, and new domains are introduced at later stages, enabling models to adapt and generalize to new unseen domains.
\begin{itemize}
    \item \textbf{Digit10:} Digit-10 dataset contains 10 digit categories in four domains: \textbf{MNIST}\cite{lecun2010mnist}, \textbf{EMNIST}\cite{cohen2017emnist}, \textbf{USPS}\cite{hull1994database}, \textbf{SVHN}\cite{netzer2011reading}.Each dataset is a digit image classification dataset of 10 classes in a specific domain, such as handwriting style.
    \begin{itemize}
        \item \textbf{MNIST:} A dataset of handwritten digits with a training set of 60,000 examples and a test set of 10,000 examples.
        \item \textbf{EMNIST:} An extended version of MNIST that includes handwritten characters (letters and digits) with a training set of 240,000 examples and a test set of 40,000 examples.
        \item \textbf{USPS:} The United States Postal Service dataset consists of handwritten digits with a training set of 7,291 examples and a test set of 2,007 examples.
        \item \textbf{SVHN:} The Street View House Numbers dataset contains images of house numbers captured from Google Street View, with a training set of 73,257 examples and a test set of 26,032 examples.
    \end{itemize}
    
    \item \textbf{Office31: } A dataset with images from three different domains: Amazon, Webcam, and DSLR. It consists of 31 object categories, with each domain having around 4,100 images.

    \item \textbf{DomainNet:} A large-scale dataset with images from six different domains: Clipart, Painting, Real, Sketch, Quickdraw, and Infograph. It contains over 0.6 million images across 345 categories.
\end{itemize}

\section{Baseline}\label{sec:baseline}
\begin{itemize}
    \item \textbf{Representative FL models:} 
    \begin{itemize}
        \item \textbf{FedAvg:} It is a representative federated learning model, which aggregates client parameters in each communication. It is a simple yet effective model for federated learning.
        \item \textbf{FedProx:} It is also a representative federated learning model, which is better at tackling heterogeneity in federated networks than FedAvg.
    \end{itemize}
    
    \item \textbf{Custom methods:}
        \begin{itemize}
        \item \textbf{Fixed:} we train the model only from the first task and evaluate it for all the coming sequence of tasks.
        \item \textbf{DANN+FL:} Here we adopt the robust adversarial-based method DANN\cite{ganin2016domain}. This baseline mainly follows the domain adaptation paradigm which is different from the incremental learning setting and are often prone to issues like catastrophic forgetting.
        \item \textbf{Shared:} Inspired by the multi-task learning scenario\cite{ruder2017overview}, we adopt all front layers before the last fully connected layer as shared layers, and use relevant different fully-connected layers to get outputs for different tasks.
    \end{itemize}
    
    \item \textbf{Models for federated class-incremental learning:}
        \begin{itemize}
        \item \textbf{FCIL:} This approach addresses the federated class-incremental learning and trains a global model by computing additional class-imbalance losses. A proxy server is introduced to reconstruct samples to help clients select the best old models for loss computation.
        \item \textbf{FedCIL:} This approach employs the ACGAN backbone to generate synthetic samples to consolidate the global model and align sample features in the output layer. Authors conduct experiments in the FCIL scenario, and here we adopt it to our FDIL setting.
    \end{itemize}  
\end{itemize}

\section{Configurations}\label{sec:config}
For local training, the batch size is 64, learning rate for our models is 0.01/0.001 for \{Office31, CIFAR10, CIFAR100\}/\{Digit10, DomainNet, Tiny-ImageNet\}. For the update of the personalized informative model, the epoch is set to $40$ for each client. For the multi-task learning structure in our approach, we treat all previous layers before the last fully-connected layer as share layers, and we use two different fully-connected layers to get outputs as the auxiliary classifier result and target classification result.
We build the Virtual Machine(VM) to simulate the experiment environment and set up different processes to simulate different clients. The VM is configured with 8 RTX4090 and 6 2.3GHz Intel Xeon CPUs. 

\onecolumn\allowdisplaybreaks

\section{Detailed Re-Fed Framework with FedAvg }\label{sec:detailed}
\begin{algorithm*}[h]
   \caption{Re-Fed for FIL with FedAvg Algorithm}
   \label{detailed}
   \SetKwInOut{Input}{Input}
    \Input{$T$: communication round;\ \ $K$: number of clients;\ \ $\eta$: learning rate;\ \ $\{\mathcal{T}^{t}\}_{t=1}^n$: distributed dataset with $n$ tasks;\ \ $w$: parameter of the model;\ \ $v_k$: personalized informative model in client $k$;\ \ $\lambda$: factor of information proportion.}
            
    Initialize the parameter $w$;
    
    \For(\tcp*[f]{When the $t$-th new task arrives}){$c=1$ {\bfseries to} $T$}{
        Server randomly selects a subset of devices $S_t$ and send $w^{t-1}$ to them;  \\      
        \For{each selected client $k \in S_t$ {\rm\bfseries in parallel}}{        
            Receive the distributed global model $w^{t-1}$ and initializess the personalized informative model $v_k^{t-1}$;\\
            $\text{Update}\ v_k^{t-1} \text{in $s$ local iterations}$ with previous local samples $\mathcal{T}^{t-1}_{k,local}$:\\
            $v_{k,s}^{t-1} = v_{k,s-1}^{t-1}-\eta\biggl(\sum_{i=1}^{M}\nabla l\left(f_{v_{k,s-1}^{t-1}}(\Tilde{x}_{k,t-1}^{(i)}),\Tilde{y}_{k,t-1}^{(i)}\right)+ q(\lambda)(v_{k,s-1}^{t-1}-w^{t-1})\biggl)$, $\ q(\lambda) =  \frac{1-\lambda}{2 \lambda},  \lambda \in (0,1).$\\
           \For{During the update of $v_k^{t-1}$}{
                Calculate the importance score for the sample $(\Tilde{x}^{(i)}_{k,t-1},\Tilde{y}^{(i)}_{k,t-1})$ after total $s$ iterations: \\
                $I(\Tilde{x}^{(i)}_{k,t-1}) = \sum_{p=1}^s\frac{G^p(\Tilde{x}^{(i)}_{k,t-1})}{p}$.
           }
        Cache previous samples with higher importance scores;\\
        Train the local model with cached samples and the new task $(\Tilde{x}^{(i)}_{k,t},\Tilde{y}^{(i)}_{k,t})$ in $s$ iterations:\\
        $w_{k,s}^{t} = w_{k,s-1}^{t}-\eta\biggl(\sum_{i=1}^{M}\nabla l\left(f_{w_{k,s-1}^{t}}(\Tilde{x}_{k,t}^{(i)}),\Tilde{y}_{k,t}^{(i)}\right).$\\
        Send the model $w_k^t$ back to the server.
        }
    The server aggregates the local models: $w^t = \sum_{k\in S_t}\frac{1}{|S_t|}w^t_k$.
    }
\end{algorithm*}

\section{Experimental Results}\label{sec:result}
In this section, we further provide more details about the experiment results on the test accuracy and communication rounds. We record the test accuracy of the {global} model at training stage of each task and the communication rounds required to achieve the corresponding performance. Then, as we use a form of ``Early-Emphasis'' to accumulate the gradient norms and calculate the sample importance scores in Re-Fed, we compare and show results with two other methods of calculation of sample importance scores.
\subsection{Detailed Results of Test Accuracy.}
Table \ref{cifar10}, \ref{cifar100}, \ref{tiny_imagenet}, \ref{digit_office} and \ref{domainnet} show the results of test accuracy on each incremental task in the \textbf{Acc} (Accuray) line, where ``$\Delta$'' denotes the improvement of our method with other baselines. Here we measure average accuracy over all tasks on each client in the \textbf{Acc} line and highlight the best test accuracy in \textbf{bold}.

\subsection{Detailed Results of Communication Round.}
Table \ref{cifar10}, \ref{cifar100}, \ref{tiny_imagenet}, \ref{digit_office} and \ref{domainnet} show the detailed results of communication round on each incremental task in the \textbf{CoR} (Communication Round) line and highlight the results of fewest number of communication rounds in \underline{underline}. 

\subsection{Different Weighting Methods for Gradient Norms.}
Table \ref{weight} shows the impact of using different methods to calculate the sample importance score with gradient norms in the update of personalized informative models. Here we adopt three methods: \textbf{Average Weighting:} we assign an equal weight to gradient norms from different iterations; \textbf{Early-Emphasis:} a higher weight to gradient norms in the early-training as adopted by Re-Fed; and \textbf{Late-Emphasis}: the sorting of samples with the sample importance score obtained by the method of \textbf{Early-Emphasis} is reversed.

\vspace*{1cm}
\begin{table*}[h]
    \centering
        \caption{Performance comparisons of various methods on CIFAR10 with 5 incremental tasks (2 new classes for each task).}
    \renewcommand\arraystretch{1.25}
      \resizebox{0.7\linewidth}{!}{
    \begin{tabular}{c  c|c c c c c| c |c}
    \toprule[1pt]
        \multicolumn{9}{c}{\textbf{CIFAR10} ($\alpha$ = 1.0)}\\
        \hline
         Method & Target & 2 & 4 & 6 & 8 & 10 & Avg &$\Delta(\uparrow)$\\
       \hline
        FedAvg&\makecell[c]{Acc \\ CoR}&\makecell[c]{92.65 \\ 142}& \makecell[c]{76.67 \\ 123}& \makecell[c]{42.90 \\ 125}& \makecell[c]{40.46 \\ 98}& \makecell[c]{26.73 \\ 119}& \makecell[c]{55.88 \\ 122}& \makecell[c]{2.57$\uparrow$ \\ 10$\uparrow$}\\

       FedProx&\makecell[c]{Acc \\ CoR}& \makecell[c]{92.39 \\ 153}& \makecell[c]{74.18 \\ 137} & \makecell[c]{39.84 \\ 141}& \makecell[c]{37.55 \\ 123} & \makecell[c]{25.87 \\ 132} & \makecell[c]{53.97 \\ 137}& \makecell[c]{4.48$\uparrow$ \\ 25$\uparrow$}\\
       \hline
       Fixed&\makecell[c]{Acc \\ CoR}& \makecell[c]{92.65 \\ 142} & \makecell[c]{62.48 \\ 0} &\makecell[c]{36.54 \\ 0} & \makecell[c]{24.20 \\ 0} & \makecell[c]{19.21 \\ 0} & \makecell[c]{47.02 \\ /} & \makecell[c]{11.43$\uparrow$ \\ /}\\

       DANN+FL&\makecell[c]{Acc \\ CoR}& \makecell[c]{93.07 \\ 151} & \makecell[c]{77.81 \\ 140} & \makecell[c]{44.32 \\ 150} & \makecell[c]{36.98 \\ 126}&\makecell[c]{24.86 \\ 145} & \makecell[c]{55.41 \\ 142}&\makecell[c]{3.04$\uparrow$ \\ 30$\uparrow$}\\
       Shared&\makecell[c]{Acc \\ CoR}&\makecell[c]{92.65 \\ 142} & \makecell[c]{76.19 \\ 117}& \makecell[c]{42.15 \\ 116}&\makecell[c]{38.24 \\ \underline{83}} &\makecell[c]{23.91 \\ 118} & \makecell[c]{54.63 \\ 115}& \makecell[c]{3.82$\uparrow$ \\ 3$\uparrow$}\\
       \hline

       FCIL&\makecell[c]{Acc \\ CoR}& \makecell[c]{92.65 \\ 142}& \makecell[c]{78.07 \\ 125}&\makecell[c]{43.66 \\ \underline{108}} & \makecell[c]{40.28 \\ 92}&\makecell[c]{25.04 \\ 121} & \makecell[c]{55.94 \\ 118}& \makecell[c]{2.51$\uparrow$ \\ 6$\uparrow$}\\
       FedCIL&\makecell[c]{Acc \\ CoR}&\makecell[c]{\textbf{94.05} \\ 148} &\makecell[c]{\textbf{80.22} \\ 150} &\makecell[c]{46.19 \\ 146} & \makecell[c]{35.50 \\ 147}&\makecell[c]{27.35 \\ 150} & \makecell[c]{56.66 \\ 148}& \makecell[c]{1.79$\uparrow$ \\ 36$\uparrow$}\\
  
       \hline
       Re-Fed &\makecell[c]{Acc \\ CoR}& \makecell[c]{92.65 \\ \underline{142}} & \makecell[c]{79.23 \\ \underline{109}} & \makecell[c]{\textbf{47.41} \\ 116}& \makecell[c]{\textbf{43.75} \\ 85} & \makecell[c]{\textbf{29.22} \\ \underline{106}} & \makecell[c]{\textbf{58.45} \\ \underline{112}} & / \\
    \bottomrule[1pt]
    \end{tabular}}
    \label{cifar10}
\end{table*}
\vspace*{0.5cm}

\begin{table*}[h]
    \centering
        \caption{Performance comparisons of various methods on CIFAR100 with 10 incremental tasks (10 new classes for each task).}
    \renewcommand\arraystretch{1.25}
      \resizebox{\linewidth}{!}{
    \begin{tabular}{c c|c c c c c c c c c c| c |c}
    \toprule[1pt]
        \multicolumn{14}{c}{\textbf{CIFAR100} ($\alpha$ = 5.0)}\\
        \hline
         Method & Target & 10 & 20 & 30 & 40 & 50 & 60 & 70 & 80 & 90 & 100 & Avg &$\Delta(\uparrow)$\\
       \hline
        FedAvg&\makecell[c]{Acc \\ CoR}& \makecell[c]{58.70 \\ 137} & \makecell[c]{43.72 \\ 121} & \makecell[c]{48.69 \\ \underline{76}} & \makecell[c]{38.28 \\ 135} & \makecell[c]{30.81 \\ 102} & \makecell[c]{26.16 \\ 143} & \makecell[c]{24.90 \\ 90} & \makecell[c]{20.72 \\ \underline{86}} & \makecell[c]{18.97 \\ 132} & \makecell[c]{17.21 \\ 75} & \makecell[c]{32.82 \\ 110} & \makecell[c]{5.57$\uparrow$ \\ 6$\uparrow$}\\
       FedProx&\makecell[c]{Acc \\ CoR}& \makecell[c]{56.51 \\ 146} & \makecell[c]{42.02 \\ 134} & \makecell[c]{48.03 \\ 112} & \makecell[c]{39.11 \\ 139} & \makecell[c]{32.33 \\ 119} & \makecell[c]{27.24 \\ 140} & \makecell[c]{26.50 \\ 125} & \makecell[c]{20.88 \\ 105} & \makecell[c]{19.67 \\ 132} & \makecell[c]{18.03 \\ 99} & \makecell[c]{33.03 \\ 125} & \makecell[c]{5.36$\uparrow$ \\ 21$\uparrow$}\\
       \hline
       Fixed&\makecell[c]{Acc \\ CoR}& \makecell[c]{58.70 \\ 137} & \makecell[c]{34.52 \\ 0} & \makecell[c]{35.09 \\ 0} & \makecell[c]{30.37 \\ 0} & \makecell[c]{27.01 \\ 0} & \makecell[c]{23.96 \\ 0} & \makecell[c]{18.18 \\ 0} & \makecell[c]{14.78 \\ 0} & \makecell[c]{11.47 \\ 0} & \makecell[c]{9.27 \\ 0} & \makecell[c]{26.34 \\ /} & \makecell[c]{12.05$\uparrow$ \\ /}\\
       DANN+FL&\makecell[c]{Acc \\ CoR}& \makecell[c]{58.82 \\ 145} & \makecell[c]{44.12 \\ 129} & \makecell[c]{46.84 \\ 123} & \makecell[c]{39.66 \\ 138} & \makecell[c]{31.54 \\ 124} & \makecell[c]{27.93\\ 134} & \makecell[c]{24.21 \\ 112} & \makecell[c]{24.03 \\ 109} & \makecell[c]{21.32 \\ 129} & \makecell[c]{19.73 \\ 121} & \makecell[c]{33.82 \\ 126} & \makecell[c]{4.57$\uparrow$ \\ 22$\uparrow$}\\
       Shared&\makecell[c]{Acc \\ CoR}& \makecell[c]{58.70 \\ 137} & \makecell[c]{42.53 \\ 117} & \makecell[c]{48.49 \\ 82} & \makecell[c]{39.10 \\ 137} & \makecell[c]{31.88 \\ 113} & \makecell[c]{27.39 \\ 137} & \makecell[c]{25.85 \\ 103} & \makecell[c]{25.74 \\ 97} & \makecell[c]{24.35 \\ 135} & \makecell[c]{18.30 \\ 89} & \makecell[c]{34.23 \\ 115} & \makecell[c]{4.16$\uparrow$ \\ 11$\uparrow$}\\
       \hline
       FCIL&\makecell[c]{Acc \\ CoR}& \makecell[c]{58.70 \\ 137} & \makecell[c]{45.65 \\ 123} & \makecell[c]{51.87 \\ 77} & \makecell[c]{42.37 \\ 134} & \makecell[c]{37.32 \\ 105} & \makecell[c]{32.01 \\ 140} & \makecell[c]{29.00 \\ 96} & \makecell[c]{28.47 \\ 88} & \makecell[c]{24.99 \\ 130} & \makecell[c]{23.02 \\ \underline{73}} & \makecell[c]{37.33 \\ 110} & \makecell[c]{1.06$\uparrow$ \\ 6$\uparrow$}\\
       FedCIL&\makecell[c]{Acc \\ CoR}& \makecell[c]{\textbf{61.20} \\ 146} & \makecell[c]{\textbf{47.05} \\ 138} & \makecell[c]{49.66 \\ 123} & \makecell[c]{38.14 \\ 131} & \makecell[c]{32.69 \\ 125} & \makecell[c]{24.11 \\ 143} & \makecell[c]{23.90 \\ 122} & \makecell[c]{23.99 \\ 129} & \makecell[c]{19.89 \\ 130} & \makecell[c]{17.98 \\ 126} & \makecell[c]{33.86 \\ 131} & \makecell[c]{4.53$\uparrow$ \\ 27$\uparrow$}\\
  
       \hline
       Re-Fed&\makecell[c]{Acc \\ CoR}& \makecell[c]{58.70 \\ \underline{137}} & \makecell[c]{43.66 \\ \underline{104}} &\makecell[c]{\textbf{53.53} \\ 80} & \makecell[c]{\textbf{40.17} \\ \underline{105}} &\makecell[c]{\textbf{38.71} \\ \underline{93}} &\makecell[c]{\textbf{35.96} \\ \underline{121}} &\makecell[c]{\textbf{31.25} \\ \underline{85}}& \makecell[c]{\textbf{28.77} \\ 105}&\makecell[c]{\textbf{27.53} \\ \underline{120}} & \makecell[c]{\textbf{25.61} \\ 87} &\makecell[c]{\textbf{38.39} \\ \underline{104}} & / \\
    \bottomrule[1pt]
    \end{tabular}}
    \label{cifar100}
\end{table*}

\newpage
\vspace*{1cm}

\begin{table*}[htbp]
    \centering
        \caption{Performance comparisons of various methods on Tiny-ImageNet with 10 incremental tasks (20 new classes for each task).}
    \renewcommand\arraystretch{1.2}
      \resizebox{\linewidth}{!}{
    \begin{tabular}{c c|c c c c c c c c c c| c |c}
    \toprule[1pt]
        \multicolumn{14}{c}{\textbf{Tiny-ImageNet} ($\alpha$ = 10.0)}\\
        \hline
         Method &Target & 20 & 40 & 60 & 80 & 100 & 120 & 140 & 160 & 180 & 200 & Avg &$\Delta(\uparrow)$\\
       \hline
        FedAvg&\makecell[c]{Acc \\ CoR}& \makecell[c]{85.80 \\ 132} & \makecell[c]{68.58 \\ 143} & \makecell[c]{57.22 \\ 139} & \makecell[c]{43.75 \\ 125} & \makecell[c]{40.52 \\ 107} & \makecell[c]{41.13 \\ \underline{97}} & \makecell[c]{34.10 \\ 128} & \makecell[c]{29.59 \\ 121} & \makecell[c]{28.40 \\ 109} & \makecell[c]{27.58 \\ 98} & \makecell[c]{45.67 \\ 120} & \makecell[c]{5$\uparrow$ \\ 7$\uparrow$}\\
       FedProx&\makecell[c]{Acc \\ CoR}& \makecell[c]{82.02 \\ 127} & \makecell[c]{66.15 \\ 140} & \makecell[c]{54.32 \\ 142} & \makecell[c]{40.57 \\ 134} & \makecell[c]{38.80 \\ 120} & \makecell[c]{38.99 \\ 113} & \makecell[c]{30.59 \\ 114} & \makecell[c]{24.12 \\ 121} & \makecell[c]{22.76 \\ \underline{110}} & \makecell[c]{21.82 \\ 108} & \makecell[c]{42.01 \\ 123} & \makecell[c]{8.66$\uparrow$ \\ 10$\uparrow$}\\
       \hline
       Fixed&\makecell[c]{Acc \\ CoR}& \makecell[c]{85.80 \\ 132} & \makecell[c]{51.07 \\ 0} & \makecell[c]{30.94 \\ 0} & \makecell[c]{28.11 \\ 0} & \makecell[c]{25.30 \\ 0} & \makecell[c]{24.26 \\ 0} & \makecell[c]{19.48 \\ 0} & \makecell[c]{17.18 \\ 0} & \makecell[c]{14.66 \\ 0} & \makecell[c]{12.34 \\ 0} & \makecell[c]{30.91 \\ /} & \makecell[c]{19.76$\uparrow$ \\ /}\\
       DANN+FL&\makecell[c]{Acc \\ CoR}& \makecell[c]{85.24 \\ 138} & \makecell[c]{68.16 \\ 140} & \makecell[c]{55.32 \\ 141} & \makecell[c]{41.11 \\ 131} & \makecell[c]{36.45 \\ 124} & \makecell[c]{35.38 \\ 126} & \makecell[c]{28.83 \\ 137} & \makecell[c]{24.54 \\ 128} & \makecell[c]{21.09 \\ 121} & \makecell[c]{20.77 \\ 123} & \makecell[c]{41.69 \\ 131} & \makecell[c]{8.98$\uparrow$ \\ 18$\uparrow$}\\
       Shared&\makecell[c]{Acc \\ CoR}& \makecell[c]{85.80 \\ 132} & \makecell[c]{67.21 \\ 135} & \makecell[c]{56.49 \\ 145} & \makecell[c]{42.05 \\ 125} & \makecell[c]{40.17 \\ 119} & \makecell[c]{37.59 \\ 127} & \makecell[c]{28.61 \\ 129} & \makecell[c]{25.90 \\ 116} & \makecell[c]{23.89 \\ 130} & \makecell[c]{22.19 \\ 125} & \makecell[c]{42.99 \\ 128} & \makecell[c]{7.68$\uparrow$ \\ 15$\uparrow$}\\
       \hline
       FCIL&\makecell[c]{Acc \\ CoR}& \makecell[c]{85.80 \\ 132} & \makecell[c]{71.94 \\ 130} & \makecell[c]{61.02 \\ 127} & \makecell[c]{50.73 \\ \underline{112}} & \makecell[c]{44.25 \\ 106} & \makecell[c]{\textbf{42.40} \\ 109} & \makecell[c]{36.96 \\ 124} & \makecell[c]{34.51 \\ 122} & \makecell[c]{31.36 \\ 121} & \makecell[c]{29.58 \\ 108} & \makecell[c]{48.86\\ 119} & \makecell[c]{1.81$\uparrow$ \\ 6$\uparrow$}\\
       FedCIL&\makecell[c]{Acc \\ CoR}& \makecell[c]{\textbf{86.43} \\ 146} & \makecell[c]{69.39 \\ 144} & \makecell[c]{58.11 \\ 137} & \makecell[c]{45.74 \\ 121} & \makecell[c]{41.02 \\ 117} & \makecell[c]{38.93 \\ 126} & \makecell[c]{31.29 \\ 132} & \makecell[c]{27.65 \\ 140} & \makecell[c]{25.17 \\ 124} & \makecell[c]{24.41 \\ 129} & \makecell[c]{44.81 \\ 132} & \makecell[c]{5.86$\uparrow$ \\ 19$\uparrow$}\\
  
       \hline
      Re-Fed&\makecell[c]{Acc \\ CoR}& \makecell[c]{85.80 \\ \underline{132}} & \makecell[c]{\textbf{72.06} \\ \underline{120}} & \makecell[c]{\textbf{65.29} \\ \underline{126}} & \makecell[c]{\textbf{52.39} \\ 121} & \makecell[c]{\textbf{45.93} \\ \underline{91}} & \makecell[c]{42.15 \\ 103} & \makecell[c]{\textbf{38.88} \\ \underline{110}} & \makecell[c]{\textbf{36.95} \\ \underline{114}} & \makecell[c]{\textbf{35.19} \\ 112} & \makecell[c]{\textbf{32.07} \\ \underline{92}} & \makecell[c]{\textbf{50.67} \\ \underline{113}} &  / \\
    \bottomrule[1pt]
    \end{tabular}}
    \label{tiny_imagenet}
\end{table*}
\vspace*{0.5cm}

\begin{table*}[h]
    \centering
        \caption{Performance comparisons of various methods on Digit10 with 4 domains and Office-31 with 3 domains.}
    \renewcommand\arraystretch{1.2}
      \resizebox{\linewidth}{!}{
    \begin{tabular}{c c|c c c c| c |c|c c c| c |c}
    \toprule[1pt]
    &&\multicolumn{6}{c|}{\textbf{Digit10} ($\alpha$ = 0.1)}&\multicolumn{5}{c}{\textbf{Office-31} ($\alpha$ = 1.0)}\\
    \hline
         Method & Target & MNIST & EMNIST & USPS & SVHN & Avg & $\Delta(\uparrow)$ & Amazon & Dlsr & Webcam & Avg &$\Delta(\uparrow)$\\
       \hline
       
       FedAvg&\makecell[c]{Acc \\ CoR}&\makecell[c]{92.82 \\ 112} &\makecell[c]{88.62 \\ 82}&\makecell[c]{84.02 \\ 96} &\makecell[c]{77.59 \\ 122} &\makecell[c]{85.76 \\103}& \makecell[c]{3.99$\uparrow$ \\ 22$\uparrow$}& \makecell[c]{58.08 \\ 144}& \makecell[c]{31.62 \\ 136} & \makecell[c]{39.25 \\ 135}& \makecell[c]{42.98 \\ 138} & \makecell[c]{8.76$\uparrow$ \\ 9$\uparrow$}\\
       FedProx&\makecell[c]{Acc \\ CoR}&\makecell[c]{93.07 \\ 114} & \makecell[c]{87.43 \\ 93}&\makecell[c]{85.67 \\ 89} &\makecell[c]{79.09 \\ 118} & \makecell[c]{86.32 \\ 103}&\makecell[c]{3.43$\uparrow$ \\ 22$\uparrow$}&\makecell[c]{58.69 \\ 145} & \makecell[c]{34.25 \\ 146} & \makecell[c]{43.01 \\ 139} &\makecell[c]{45.32 \\ 143} & \makecell[c]{6.42$\uparrow$ \\ 14$\uparrow$}\\
       \hline
       Fixed&\makecell[c]{Acc \\ CoR}&\makecell[c]{92.82 \\ 112}&\makecell[c]{85.35 \\ 0}&\makecell[c]{82.11 \\ 0}& \makecell[c]{71.26 \\ 0}&\makecell[c]{82.48 \\ /} & \makecell[c]{7.27$\uparrow$ \\ /}&\makecell[c]{58.08 \\ 144} & \makecell[c]{24.56 \\ 0} & \makecell[c]{37.44 \\ 0} & \makecell[c]{40.03 \\ /} & \makecell[c]{11.71$\uparrow$ \\ /}\\
       DANN+FL&\makecell[c]{Acc \\ CoR}&\makecell[c]{\textbf{96.07} \\ 132} &\makecell[c]{87.30 \\ 107}&\makecell[c]{82.81 \\ 116}&\makecell[c]{76.44 \\ 129}&\makecell[c]{85.66 \\ 120}& \makecell[c]{4.09$\uparrow$ \\ 39$\uparrow$}&\makecell[c]{\textbf{59.95} \\ 149}&\makecell[c]{42.21 \\ 144}& \makecell[c]{45.21 \\ 141}& \makecell[c]{49.12 \\ 145} & \makecell[c]{2.62$\uparrow$ \\ 16$\uparrow$}\\
       Shared&\makecell[c]{Acc \\ CoR}&\makecell[c]{92.82 \\ 112}&\makecell[c]{82.10 \\ 76}&\makecell[c]{80.36 \\ 84} &\makecell[c]{74.77 \\ 103}&\makecell[c]{82.51 \\ 93}& \makecell[c]{7.24$\uparrow$ \\ 12$\uparrow$}&\makecell[c]{58.08 \\ 144}&\makecell[c]{35.33 \\ \underline{122}}&\makecell[c]{37.55 \\ 124}&\makecell[c]{43.65 \\ 130}& \makecell[c]{8.09$\uparrow$ \\ 1$\uparrow$}\\
       \hline
       FCIL&\makecell[c]{Acc \\ CoR}&\makecell[c]{92.82 \\ 112} &\makecell[c]{88.62 \\ 82}&\makecell[c]{84.02 \\ 96} &\makecell[c]{77.59 \\ 122} &\makecell[c]{85.76 \\103}& \makecell[c]{3.99$\uparrow$ \\ 22$\uparrow$}& \makecell[c]{58.08 \\ 144}& \makecell[c]{31.62 \\ 136} & \makecell[c]{39.25 \\ 135}& \makecell[c]{42.98 \\ 138} & \makecell[c]{8.76$\uparrow$ \\ 9$\uparrow$}\\
       FedCIL&\makecell[c]{Acc \\ CoR}&\makecell[c]{94.61 \\ 118} &\makecell[c]{90.24 \\ 86} &\makecell[c]{87.55 \\ 92} &\makecell[c]{83.85 \\ 125} & \makecell[c]{89.06 \\ 105}&\makecell[c]{0.69$\uparrow$ \\ 24$\uparrow$}&\makecell[c]{59.37 \\ 146} & \makecell[c]{45.91 \\ 139} &\makecell[c]{46.26 \\ 148} &\makecell[c]{50.51 \\ 144} & \makecell[c]{1.23$\uparrow$ \\ 15$\uparrow$}\\

       \hline
       Re-Fed&\makecell[c]{Acc \\ CoR}& \makecell[c]{92.82 \\ \underline{112}} &\makecell[c]{\textbf{91.64} \\ \underline{68}}& \makecell[c]{\textbf{88.57} \\ \underline{73}}&\makecell[c]{\textbf{85.96} \\ \underline{71}}&\makecell[c]{\textbf{89.75} \\ \underline{81}} & / & \makecell[c]{58.08 \\ \underline{144}} &\makecell[c]{\textbf{47.07} \\ 125}&\makecell[c]{\textbf{50.80} \\ \underline{118}}&\makecell[c]{\textbf{51.74} \\ \underline{129}}&/ \\

    \bottomrule[1pt]
    \end{tabular}}
    \label{digit_office}
\end{table*}

\newpage
\vspace*{1cm}

\begin{table*}[htbp]
    \centering
        \caption{Performance comparisons of various methods on DomainNet with 6 domains.}
    \renewcommand\arraystretch{1.2}
      \resizebox{0.8\linewidth}{!}{
    \begin{tabular}{c c|c c c c c c | c |c}
    \toprule[1pt]
        &\multicolumn{9}{c}{\textbf{DomainNet} ($\alpha$ = 10)}\\
        \hline
         Method &Target&Clipart  & Infograph  & Painting&Quickdraw&Real&Sketch&Avg &$\Delta(\uparrow)$\\
       \hline
        FedAvg&\makecell[c]{Acc \\ CoR}&\makecell[c]{52.07 \\ 141}&\makecell[c]{36.22 \\ 128} & \makecell[c]{45.09 \\ 97}& \makecell[c]{46.59 \\ 108}& \makecell[c]{49.36 \\ 136} &\makecell[c]{51.73 \\ 115}&\makecell[c]{46.84 \\ 121}&\makecell[c]{3.39$\uparrow$ \\ 11$\uparrow$}\\
       FedProx&\makecell[c]{Acc \\ CoR}&\makecell[c]{50.31 \\ \underline{136}}&\makecell[c]{33.64 \\ 131} & \makecell[c]{41.77 \\ 115}& \makecell[c]{45.04 \\ 130}& \makecell[c]{47.44 \\ 137} &\makecell[c]{49.12 \\ 116}&\makecell[c]{44.55 \\ 128}& \makecell[c]{5.68$\uparrow$ \\ 1$\uparrow$}\\
       \hline
       Fixed&\makecell[c]{Acc \\ CoR}&\makecell[c]{52.07 \\ 141}&\makecell[c]{29.58 \\ 0}&\makecell[c]{32.24 \\ 0} &\makecell[c]{38.91 \\ 0}&\makecell[c]{40.09 \\ 0}&\makecell[c]{46.30 \\ 0}&\makecell[c]{39.87 \\ /} &\makecell[c]{10.36$\uparrow$ \\ /}\\
       DANN+FL&\makecell[c]{Acc \\ CoR}&\makecell[c]{\textbf{55.66} \\ 142}&\makecell[c]{36.44 \\ 126}  &\makecell[c]{42.02 \\ 109} & \makecell[c]{38.84 \\ 112} & \makecell[c]{45.89 \\ 137}&\makecell[c]{50.01 \\ 121}&\makecell[c]{44.81 \\ 125}&\makecell[c]{5.42$\uparrow$ \\ 15$\uparrow$}\\
       Shared&\makecell[c]{Acc \\ CoR}& \makecell[c]{52.07 \\ 141}&\makecell[c]{35.22 \\ 113}&\makecell[c]{37.83 \\ 98}& \makecell[c]{35.19 \\ 125}&\makecell[c]{40.52 \\ 120}&\makecell[c]{41.76 \\ 96}&\makecell[c]{40.43 \\ 116}&\makecell[c]{9.80$\uparrow$ \\ 6$\uparrow$}\\
       \hline
       FCIL&\makecell[c]{Acc \\ CoR}&\makecell[c]{52.07 \\ 141}&\makecell[c]{36.22 \\ 128} & \makecell[c]{45.09 \\ 97}& \makecell[c]{46.59 \\ \underline{108}}& \makecell[c]{49.36 \\ 136} &\makecell[c]{51.73 \\ 115}&\makecell[c]{46.84 \\ 121}&\makecell[c]{3.39$\uparrow$ \\ 11$\uparrow$}\\
       FedCIL&\makecell[c]{Acc \\ CoR}&\makecell[c]{54.52 \\ 148}&\makecell[c]{38.98 \\ 136} & \makecell[c]{40.45 \\ 128}& \makecell[c]{41.77 \\ 112}& \makecell[c]{45.09 \\ 142} &\makecell[c]{47.28 \\ 125} & \makecell[c]{44.68 \\ 132}& \makecell[c]{5.55$\uparrow$ \\ 22$\uparrow$}\\
  
       \hline
       Re-Fed&\makecell[c]{Acc \\ CoR}&\makecell[c]{52.07 \\ 141}&\makecell[c]{\textbf{42.26} \\ \underline{103}} & \makecell[c]{\textbf{48.11} \\ \underline{97}}& \makecell[c]{\textbf{48.98} \\ 109}& \makecell[c]{\textbf{53.34} \\ \underline{118}} &\makecell[c]{\textbf{56.66} \\ \underline{91}}&\makecell[c]{\textbf{50.23} \\ \underline{110}} & / \\
    
    \bottomrule[1pt]
    \end{tabular}}
    \label{domainnet}
\end{table*}
\vspace*{0.5cm}

\begin{table*}[htbp]
    \centering
        \caption{Performance comparisons of three weighting methods for gradient norms in two incremental scenarios.}
    \renewcommand\arraystretch{1.2}
      \resizebox{0.8\linewidth}{!}{
    \begin{tabular}{c |c c c | c c c }
    \toprule[1pt]
         \multirow{2}{*}{Dataset}  & \multicolumn{3}{c|}{Class-Incremental Scenario} & \multicolumn{3}{c}{Domain-Incremental Scenario} \\
       \cline{2-7}
        & CIFAR10 & CIFAR100 & Tiny-ImageNet & Digit10 & Office31 & DomainNet\\
       \hline
       
       Early-Emphasis & \textbf{29.22} & \textbf{25.61} & \textbf{32.07} & \textbf{85.96} & \textbf{50.80} & \textbf{56.66} \\
        Average-Weighting & 28.73 & 24.88 & 30.42 & 85.71 & 48.95 & 56.04 \\
        Late-Emphasis & 26.57 & 22.18 & 28.08 & 84.36 & 47.29 & 53.90 \\
    \bottomrule[1pt]
    \end{tabular}}
    \label{weight}
\end{table*}

\newpage
\section{Analysis of the Federated Incremental-Learning Framework: Re-Fed}
In this section, we prove the convergence of personalized informative models. To simplify the notation, here we conduct an analysis on a fixed task while the convergence does not depend on the IL setting. We first define following standard assumptions.

\noindent\textbf{Assumption 1}\ ($\textit{L}_2$ Distance.) The $\textit{L}_2$ distance between the optimal local models $\hat{w_k}$:$ = \mathop{\arg\min}\limits_{w_k}\{f(w_k)\}$ and the optimal global model $\hat{w}$:$ = \mathop{\arg\min}\limits_{w}\{\frac{1}{K}\sum_{k=1}^K\nabla f(w_k)\}$ is bounded by:
\begin{align}
    ||\hat{w_k}-\hat{w}|| \leq M, \ \forall k\in [ K ].
\end{align}


\noindent\textbf{Assumption 2}\ (Gradient Variance.) The variance of stochastic gradients is finite and bounded at all clients by:
\begin{align}
     \mathbb{E}\Big[||\nabla f(\hat{w_k})||^2\Big] \leq \sigma^2, \ \forall k\in [ K ].
\end{align}

\noindent\textbf{Assumption 3}\ (Strong Convexity.) There exists $\mu_k \in \mathbb{R}_+$ and a unique solution $\hat{w_k}$:
\begin{align}
    f(w_k)-f(\hat{w_k}) \geq \langle \nabla f(\hat{w_k}), \hat{w_k}-w_k \rangle + \frac{\mu_k}{2}||w_k-\hat{w_k}||^2.
\end{align}

\subsection{Proof of Theorem 3.1}\label{sec:thm2}


\noindent\textbf{Definition 1}\ (Personalized Informative Model Formulation.)
Denote the objective of personalized informative model $v_k$ on client $k$ while $f(\cdot)$ is strongly convex as:
\begin{align}\label{PIL_object}
    \begin{split}
       &\hat{v_k}(\lambda) := \mathop{\arg\min}\limits_{v_k}\Big\{f(v_k) + \frac{q(\lambda)}{2}||v_k-\hat{w}||^2\Big\}\\
        &q(\lambda) := \frac{1-\lambda}{2 \lambda}, \ \lambda \in (0,1)\\
    \end{split}
\end{align}
where $\hat{w}$ denotes the global model.\\

\noindent\textbf{Lemma 1}\ (Proportion of Global and Local Information.)
\textit{For all $\lambda \in (0,1)$ and $\lambda \rightarrow f(\lambda_k)$ is non-increasing:}
\begin{align}
    \begin{split}
        &\frac{\partial  f(\hat{v_k}(\lambda))}{\partial \lambda} \leq 0 \\
        &\frac{\partial ||\hat{v_k}(\lambda)-\hat{w}||^2}{\partial \lambda} \geq 0 .
    \end{split}
\end{align}
Then, for $k \in [K]$, we can get:
\begin{align}
    \lim_{\lambda \rightarrow 0}\hat{v_k}(\lambda) := \hat{w}.
\end{align}

\noindent \textit{Proof.} The proof here directly follows the proof in Theorem 3.1 \cite{DBLP:journals/corr/abs-2002-05516}.
As $\lambda$ declines and $q(\lambda)$ grows, the objective of Eq. ~\ref{PIL_object} tends to optimize $||v_k-\hat{w}||^2$ and increase the local empirical training loss $ f(v_k)$, leading to the convergence on the global model. Hence we can modify the $\lambda$ value to adjust the optimization direction of our model $v_k$ thus the dominance of local and global model information.

\noindent\textbf{Theorem 3.1}\ (Personalized Informative Model.) \textit{
Assuming the global model $w^t$ converges to the optimal model $\hat{w}$ with $g(t)$ for any client $k \in [K]$ at each communication round $t$: $ \mathbb{E}\Big[||w^t-\hat{w}||^2\Big] \leq g(t)$ and $\lim_{t\rightarrow\infty}g(t)=0$, then there exists a constant $C < \infty$ such that the personalized informative model $v_k^t$ can converge to the optimal model $\hat{v_k}$ with $Cg(t)$.}

\noindent \textit{Proof.} Here we first introduce the Lemma 2 here proved by \cite{DBLP:journals/corr/abs-2012-04221} Lemma 13.

\noindent\textbf{Lemma 2}\ (\cite{DBLP:journals/corr/abs-2012-04221} Lemma 13.)
Under assumptions above, $f(v_k)$ is $\mu_k$-strongly convex at each communication round $t$, we have:
\begin{align}
    \begin{split}
        \mathbb{E}\Big[||v_k^{t+1}-\hat{v_k}||^2\Big] 
        \leq &\left(1-\eta(\mu_k+q(\lambda))\right) \mathbb{E}\Big[||v_k^{t}-\hat{v_k}||^2\Big]+ \eta^2\left(\sigma+q(\lambda)(M+\frac{\sigma}{\mu_k})\right)^2+\eta^2q(\lambda)^2\mathbb{E}\Big[||w^{t}-\hat{w}||^2\Big]\\
        &+ 2\eta^2q(\lambda)\left(\sigma+q(\lambda)(M+\frac{\sigma}{\mu_k})\right)\sqrt{\mathbb{E}\Big[||w^{t}-\hat{w}||^2\Big]} + 2\eta q(\lambda)\sqrt{\mathbb{E}\Big[||v_k^{t}-\hat{v_k}||^2\Big]\mathbb{E}\Big[||w^{t}-\hat{w}||^2\Big]}.
    \end{split}
\end{align}
Assume $g(t+1) \le g(t)$ and let positive number $A$ be chosen such that $A({g(t)-g(t+1)})\le {g^2(t)}$, and we arrive at  $(1-\frac{g(t)}{A})g(t) \le g(t+1)$. 
Then, we prove the \textbf{Theorem 3.2} by induction. Assuming that $\mathbb{E}\Big[||v^t_k-\hat{v_k}||^2\Big] \leq Cg(t)$ where $C > 0$ and $C \geq \frac{\mathbb{E}\Big[||v^0_k-\hat{v_k}||^2\Big]}{g(0)}$, the learning rate $\eta = \frac{2g(t)}{A(\mu_k+q(\lambda))}$,  here we can continue with \textbf{Lemma 2}:
\begin{align}\label{thm2:16}
    \begin{split}
        \mathbb{E}\Big[||v_k^{t+1}-\hat{v_k}||^2\Big] \leq &(1-\frac{2g(t)}{A})Cg(t)+ \frac{4q(\lambda)\sqrt{C}g(t)}{A(\mu_k+q(\lambda))}\\
        &+ \frac{4g(t)^2}{A^2(\mu_k+q(\lambda))^2}\left((\sigma+q(\lambda)(M+\frac{\sigma}{\mu_k}))^2+q(\lambda)^2g(t)+2q(\lambda)\sqrt{g(t)}(\sigma+q(\lambda)(M+\frac{\sigma}{\mu_k}))\right).
    \end{split}
\end{align}
Therefore, if we let $C$ = $\max\{ \frac{\mathbb{E}\Big[||v^0_k-\hat{v_k}||^2\Big]}{g(0)}, 16, \frac{4\left((\sigma+q(\lambda)(M+\frac{\sigma}{\mu_k}))^2+q(\lambda)^2g(t)+2q(\lambda)\sqrt{g(t)}(\sigma+q(\lambda)(M+\frac{\sigma}{\mu_k}))\right)}{A(\mu_k+q(\lambda))^2(1-\frac{1}{(1+\frac{\mu_k}{q(\lambda)})})} \}  $, then we have:
\begin{align}\label{thm2:17}
    \begin{split}
    &\frac{4q(\lambda)\sqrt{C}g(t)^2}{A(\mu_k+q(\lambda))} + \frac{4g(t)^2}{A^2(\mu_k+q(\lambda))^2}\left((\sigma+q(\lambda)(M+\frac{\sigma}{\mu_k}))^2+q(\lambda)^2g(t)+2q(\lambda)\sqrt{g(t)}(\sigma+q(\lambda)(M+\frac{\sigma}{\mu_k}))\right) \leq \\
    & \frac{q(\lambda)Cg(t)^2}{A(\mu_k+q(\lambda))} + \frac{4g(t)^2}{A^2(\mu_k+q(\lambda))^2}\left((\sigma+q(\lambda)(M+\frac{\sigma}{\mu_k}))^2+q(\lambda)^2g(t)+2q(\lambda)\sqrt{g(t)}(\sigma+q(\lambda)(M+\frac{\sigma}{\mu_k}))\right) = \\
    & \frac{Cg(t)^2}{A} \cdot \frac{1}{(1+\frac{\mu_k}{q(\lambda)})} + \frac{4g(t)^2}{A^2(\mu_k+q(\lambda))^2}\left((\sigma+q(\lambda)(M+\frac{\sigma}{\mu_k}))^2+q(\lambda)^2g(t)+2q(\lambda)\sqrt{g(t)}(\sigma+q(\lambda)(M+\frac{\sigma}{\mu_k}))\right) \leq\\
    & \frac{Cg(t)^2}{A} \cdot \frac{1}{(1+\frac{\mu_k}{q(\lambda)})} + \frac{g(t)^2}{A^2} \cdot CA\left(1-\frac{1}{(1+\frac{\mu_k}{q(\lambda)})}\right) = \frac{Cg(t)^2}{A}.\\
    \end{split}
\end{align}
The first inequality uses the fact that $16\le C$ and consequently $4\sqrt{C}\le C$. The second inequality results from the definition of $C$ as $\frac{4\left((\sigma+q(\lambda)(M+\frac{\sigma}{\mu_k}))^2+q(\lambda)^2g(t)+2q(\lambda)\sqrt{g(t)}(\sigma+q(\lambda)(M+\frac{\sigma}{\mu_k}))\right)}{A(\mu_k+q(\lambda))^2} \} \le C(1-\frac{1}{(1+\frac{\mu_k}{q(\lambda)})})$.
Hence, combining the results of \ref{thm2:16} and \ref{thm2:17} yields
\begin{align}
    \begin{split}
        \mathbb{E}\Big[||v_k^{t+1}-\hat{v_k}||^2\Big] &\leq (1-\frac{2g(t)}{A})Cg(t)+\frac{Cg(t)^2}{A}\\[0.8em]
        &= (1-\frac{g(t)}{A})Cg(t)\\[0.8em]
        &\leq Cg(t+1),
    \end{split}
\end{align}
and we have the desired result.

\end{document}